\documentclass{IEEEtran}
\usepackage{amsmath,amssymb,amsfonts,amsthm,bbding,textcomp} 
\usepackage{MnSymbol}
\usepackage{booktabs,multirow,threeparttable,subfigure,colortbl} 
\usepackage{enumitem}
\usepackage{graphicx}
\usepackage[square,numbers,sort&compress]{natbib}
\usepackage[svgnames]{xcolor}
\usepackage[colorlinks=false, anchorcolor=SteelBlue, linkcolor=SteelBlue, citecolor=SteelBlue]{hyperref}
\newcommand{\eqaref}[1]{(\ref{#1})}
\newcommand{\figref}[1]{Fig.\ \ref{#1}}
\newcommand{\tabref}[1]{Table\ \ref{#1}}
\newcommand{\thmref}[1]{Theorem\ \ref{#1}}
\newcommand{\lemref}[1]{Lemma\ \ref{#1}}
\newcommand{\sig}{\rlap{$^*$}}
\newcommand{\dd}{\mathop{d}}

\usepackage{algorithm}
\usepackage[noend]{algpseudocode}
\definecolor{abl}{HTML}{cda380}
\newtheorem{thm}{Theorem}
\newtheorem{lem}[thm]{Lemma}
\newtheorem{cor}[thm]{Corollary}

\begin{document}
\title{Entire Space Counterfactual Learning for Reliable Content Recommendations}

\author{Hao Wang, Zhichao Chen, Zhaoran Liu, Haozhe Li, Degui Yang, Xinggao Liu, Haoxuan Li, \IEEEmembership{Member,~IEEE}
\thanks{Hao Wang, Zhichao Chen, Zhaoran Liu, Haozhe Li, Xinggao Liu are with the State Key Laboratory of Industrial Control Technology, College of Control Science and Engineering, Zhejiang University, Hangzhou 310027, China (e-mail: Ho-ward@outlook.com; 12032042@zju.edu.cn; 22032057@zju.edu.cn; lihaozhe@zju.edu.cn; lxg@zju.edu.cn).
Degui Yang is with the School of Automation, Central South University, Changsha, 410000, China (email: degui.yang@csu.edu.cn).
Haoxuan Li is with the Center for Data Science, Peking University, Beijing, 100871, China (email: hxli@stu.pku.edu.cn).
}
\thanks{Haoxuan Li is the corresponding author.}
}

\markboth{IEEE TRANSACTIONS ON Information Forensics and Security}%
{How to Use the IEEEtran \LaTeX \ Templates}

\newcommand{\ie}{\textit{i}.\textit{e}., }
\newcommand{\eg}{\textit{e}.\textit{g}., }
\maketitle
\begin{abstract}
 Post-click conversion rate (CVR) estimation is a fundamental task in developing effective recommender systems, yet it faces challenges from data sparsity and sample selection bias. To handle both challenges, the entire space multitask models are employed to decompose the user behavior track into a sequence of exposure $\rightarrow$ click $\rightarrow$ conversion, constructing surrogate learning tasks for CVR estimation. However, these methods suffer from two significant defects: (1) intrinsic estimation bias (IEB), where the CVR estimates are higher than the actual values;  (2) false independence prior (FIP), where the causal relationship between clicks and subsequent conversions is potentially overlooked.  To overcome these limitations, we develop a model-agnostic framework, namely Entire Space Counterfactual Multitask Model (ESCM$^2$), which incorporates a counterfactual risk minimizer within the ESMM framework to regularize CVR estimation. Experiments conducted on large-scale industrial recommendation datasets and an online industrial recommendation service demonstrate that ESCM$^2$ effectively mitigates IEB and FIP defects and substantially enhances recommendation performance.

\end{abstract}

\begin{IEEEkeywords}
Debiased Recommendation; Multitask Learning; Conversion Rate Estimation.
\end{IEEEkeywords}

\section{Introduction}
\begin{figure}
\centering
\includegraphics[width=1\linewidth]{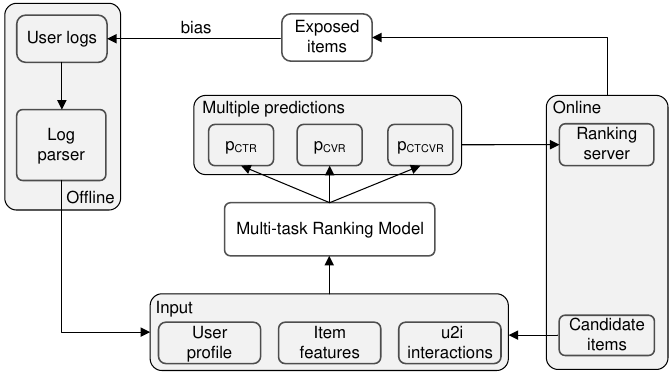}
\caption{Overview of a two-stage recommender in industry, which mainly involves the offline and online phases.\label{fig:recsys}}
\end{figure}

\IEEEPARstart{R}{ecommendation} systems play an essential role in customizing content delivery across various industries such as e-commerce~\cite{esmm}, advertising~\cite{zhou2018deep}, and social media~\cite{social}, serving as a cornerstone in information management and dissemination~\cite{DBLP:journals/tifs/LaiZFZZ24,DBLP:journals/tifs/ZhangCZWL21}. 
These systems typically operate through a two-phase pipeline, as illustrated in \figref{fig:recsys}, involving an offline phase and an online phase. In the offline phase, user profiles, item attributes, and user-item interactions are extracted from logs to train a multitask ranking model. In the online phase, this model ranks candidate items based on criteria of interest such as click-through rate (CTR), post-click conversion rate (CVR), and click-through and conversion rate (CTCVR), and exposes top items to users to meet their preferences. The model is continuously refined based on real-time user feedback~\cite{esmm}.

User behavior in these systems often follows a trajectory of exposure $\to$ click $\to$ conversion~\cite{esmm}. In this context, CTR, CVR, and CTCVR respectively quantify the transition probabilities from exposure to click, click to conversion, and exposure to conversion. Advances in feature interaction and deep learning have established CTR estimation as a good practice. However, click-through feedback is often interfered by unexpected factors such as clickbait, which distorts genuine user preferences~\cite{clickbait2}. In contrast, CVR, which indicates post-click user behavior, offers a more accurate reflection of user preferences, attracting focused attention within the recommendation community~\cite{mtlips,escm,li2024removing}.

A na\"ive yet common approach~\cite{treatment} to obtaining CVR estimators is to train models solely with samples where click happens. Although conversion labels are fully observable~\cite{treatment} with clicked samples, this na\"ive approach introduces two major problems: sample selection bias and data sparsity. Sample selection bias arises because the training dataset consists only of clicked samples, whereas the inference needs to consider all exposures~\cite{mtlips}.
Since samples with lower CVR are more likely to be excluded from the click (training) space \cite{marlin2012collaborative,mtlips}, training data is missing not at random~\cite{treatment}, which leads to a distribution shift between the training and inference spaces~\cite{wang2024optimal}.
Data sparsity arises due to low click-through rates (e.g., 4\% in the Ali-CPP dataset and 3.8\% in our industrial dataset), which restricts the data availability for training CVR estimators~\cite{esmm}.
These two problems undermine the reliability of recommendation systems by hampering their generalization to unseen samples.

\begin{figure}
\centering
\includegraphics[width=\linewidth]{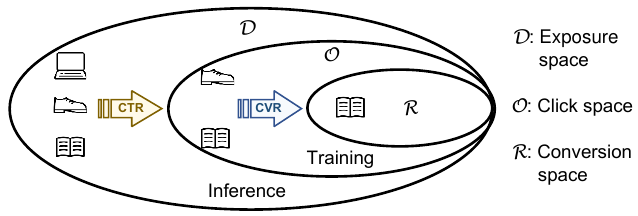}
\caption{Overview of the CVR estimation task in recommendation systems, wherein online inference is executed on all exposed samples, while training is exclusively carried out on clicked samples, leading to data sparsity and sample selection bias.\label{fig:yangcong}}
\end{figure}

To handle the challenge of data sparsity and sample selection bias, the entire space multitask model (ESMM)~\cite{esmm} avoids training CVR estimator directly by employing a multitask approach that simultaneously optimizes CTCVR and CTR objectives~\cite{esmm}.
Since training of both CTCVR and CTR objectives can utilize all exposure samples, this strategy alleviates data sparsity and enhances performance in practice~\cite{esm2}. However, its reliability in CVR estimation has been questioned due to the lack of unbiasedness guarantee~\cite{mtlips} and overly simplistic dependency assumptions. In this study, we formalize these concerns as two defects of ESMM:
\begin{itemize}[leftmargin=*]
    \item \textbf{Intrinsic Estimation Bias (IEB)}: the CVR estimates are biased from true values.
    \item \textbf{False Independence Prior (FIP)}: the CTR and CVR estimates are susceptible to inappropriate assumptions of conditional independence, ignoring the causal relationship from click to conversion.
\end{itemize}

To address these defects, we propose the Entire Space Counterfactual Multitask Model (ESCM$^2$), a model-agnostic framework that incorporates counterfactual regularizers for CVR estimation. Both theoretical and empirical evaluations demonstrate that our regularizers effectively mitigate the IEB and FIP defects. Our main contributions are as follows:
\begin{itemize}[leftmargin=*]
    \item We identify IEB and FIP as critical defects in ESMM, supported by experimental results and mathematical analysis.
    \item We develop ESCM$^2$, integrating counterfactual regularizers within ESMM to enhance performance. We provide theoretical justifications and empirical validations demonstrating its effectiveness in mitigating IEB and FIP defects.
    \item We conduct extensive evaluations using industrial recommendation datasets to validate the efficacy of ESCM$^2$. Additionally, we implement ESCM$^2$ on our online recommendation platform, where it achieves substantial profit increases\footnote{Building on our conference work~\cite{escm}, we detail the statistical properties of the regularizers, showing that both IPS and DR regularizers effectively handle IEB and FIP. We also demonstrate that IPS is a specialized importance sampling method.}.
\end{itemize}

The remaining sections are structured as follows:
Section~\ref{sec:pre} provides preliminaries for understanding the technical details in this work;
Section~\ref{sec:analyze} formulates the IEB and FIP defects with ESMM;
Section~\ref{sec:method} introduces the ESCM$^2$ framework for training recommendation models, which enhances ESMM by incorporating the proposed counterfactual regularizers to handle the IEB and FIP defects;
Section~\ref{sec:exp} presents real-world case studies to demonstrate the efficacy of ESCM$^2$;
Section~\ref{sec:rw} provides a brief overview of related works; 
Section~\ref{sec:conclusion} summarizes the conclusions and outlines open questions.

\section{Preliminaries}\label{sec:pre}
\subsection{Notations}
In this paper, uppercase letters, \eg $O$, represent random variables; lowercase letters, \eg $o$, represent the associated specific values; calligraphic letters such as $\mathcal{O}$ denote sample spaces; $\mathbb{P}(\cdot)$, $\mathbb{E}(\cdot)$, $\mathbb{V}(\cdot)$ represent probability distribution, expectation and variance, respectively.

\subsection{Problem Statement}\label{sec:state}
Denote $\mathcal{U}=\{u_1, u_2,...,u_m\}$ and $\mathcal{I}=\{i_1, i_2,...,i_n\}$  as the respective sets of users and items in the exposure space. Let $\mathcal{D}=\mathcal{U}\times\mathcal{I}$ be the set of user-item intersections in the exposure space. Let $\mathbf{O}\in\{0,1\}^{m\times n}$ be the click indicators where $o_{u,i}\in\{0,1\}$ indicates whether the user $u$ clicks the item $i$; $\mathbf{R}\in\{0,1\}^{m\times n}$ be the conversion labels where $r_{u,i}\in\{0,1\}$ indicates whether the user $u$ purchases the item $i$. 

If all entries $r_{u,i}\in\mathbf{R}$ are observable, the ideal learning objective for constructing CVR estimator is expressed as
\begin{equation}\label{eq:ideal}
    \mathcal{P}:=\mathbb{E}_{(u,i)\in\mathcal{D}}\left[\delta\left(r_{u,i}, \hat{r}_{u,i}\right)\right],
\end{equation}
where $\hat{r}_{u,i}$ denotes the estimate of $r_{u,i}$, $\delta$ measures the estimation error and can be specified as any classification loss function, $\delta\left(r_{u,i}, \hat{r}_{u,i}\right)$ is the estimation error of CVR for a specific user and item. Following existing works~\cite{mtlips,esmm}, we utilize binary cross-entropy as the loss measure:
\begin{equation}\label{eq:cvrerrormeasure}
\begin{aligned}
    \varepsilon_{u,i}:&=\delta\left(r_{u,i}, \hat{r}_{u,i}\right)\\&=-r_{u,i}\log\hat{r}_{u,i}-(1-r_{u,i})\log(1-\hat{r}_{u,i}),
\end{aligned}
\end{equation}
where we abbreviate the CVR estimation error $\delta\left(r_{u,i}, \hat{r}_{u,i}\right)$ as $\varepsilon_{u,i}$. However, the ideal objective \eqref{eq:ideal} is incomputable since $r_{u,i}$ is unobservable for samples outside the click space $\mathcal{O}$. A naive yet common shortcut is to estimate the learning objective using clicked samples in $\mathcal{O}$:
\begin{equation}
\label{eq:naive}
    \mathcal{L}_\mathrm{naive}:=\mathbb{E}_{(u,i)\in\mathcal{O}}(\varepsilon_{u,i})=\frac{1}{|\mathcal{O}|}\sum_{(u,i)\in\mathcal{D}}(o_{u,i}\varepsilon_{u,i}),
\end{equation}
where $|\mathcal{O}|=\sum_{(u,i)\in\mathcal{D}}(o_{u,i})$. Nonetheless, it has been shown that \eqref{eq:naive} is a biased estimation of the ideal objective~\cite{treatment,dr} due to selection bias, \ie $\mathbb{E}_{O}[\mathcal{L}_\mathrm{naive}]\neq\mathcal{P}$.
\subsection{Entire Space Multitask Model Approach}
The entire space multitask model (ESMM)~\cite{esmm} is prevalent in recommendation scenarios where CVR estimation plays critical roles.
To bypass data sparsity and sample selection bias in training CVR estimators, ESMM avoids direct training of the CVR estimator. Instead, it employs a multitask approach to optimize two independent learning objectives for CTCVR and CTR~\cite{esmm}. Specifically, according to the sequential user behavior track in \figref{fig:yangcong}, CVR can be represented as the quotient of CTCVR and CTR:
\begin{equation*}
\label{eq:esmmDecom}
    \underbrace{\mathbb{P}(r_{u,i}=1\mid o_{u,i}=1)}_{\mathrm{CVR}}=\frac{\overbrace{\mathbb{P}(r_{u,i}=1,o_{u,i}=1)}^{\mathrm{CTCVR}}}{\underbrace{\mathbb{P}(o_{u,i}=1)}_{\mathrm{CTR}}}.
\end{equation*}
On this basis, ESMM constructs two predictive arms to estimate respective CTR and CVR as $\hat{o}_{u,i}$ and $\hat{r}_{u,i}$, and multiplies them to acquire the estimate of CTCVR. During training, ESMM minimizes the empirical risk for CTR and CTCVR estimations as follows:
\begin{equation}\label{eq:esmmLoss}
\begin{aligned}
\mathcal{L}_\mathrm{CTR} &:= \mathbb{E}_{(u,i)\in\mathcal{D}}\left[\delta\left(o_{u,i},\hat{o}_{u,i}\right)\right]\\
\mathcal{L}_\mathrm{CTCVR} &:= \mathbb{E}_{(u,i)\in\mathcal{D}}\left[\delta\left(o_{u,i}*r_{u,i},\hat{o}_{u,i}*\hat{r}_{u,i}\right)\right].
\end{aligned}
\end{equation}
In the inference phase, the output from the CVR arm provides the CVR estimates.
Since both CTCVR and CTR objectives can utilize all exposure samples, ESMM alleviates data sparsity and enhances performance in practice~\cite{esm2}.
Moreover, ESMM seemingly circumvents sample selection bias by avoiding direct training of the CVR estimator. However, as elaborated in Section~\ref{sec:ieb}, it remains vulnerable to selection bias, formalized as the IEB defect in this work.
To adequately handle the sample selection bias, a regularization term tailored for CVR estimation seems imperative, aiming to estimate the unbiased learning objective of CVR estimator $\mathcal{P}$ with biased click data (where conversion labels are available).

\subsection{Causal Recommendation Approach}\label{sec:causalmethod}
To estimate the ideal learning objective of CVR estimator, causal inference techniques have received special attention by the recommendation community~\cite{li2022multiple,li2022stabilized}.
The core of these methods is to weight the samples in the click space to approximate the sample distribution in the exposure space.  A prominent technique is the inverse propensity score (IPS) \cite{treatment}, which weights CVR estimation errors ${\varepsilon_{u,i}}$ of clicked samples using the inverse propensity score:
\begin{equation}
\label{eq:ips}
    \mathcal{L}_\mathrm{IPS}
    :=\mathbb{E}_{(u, i) \in \mathcal{D}}\left[\frac{o_{u,i}\varepsilon_{u,i}}{q_{u,i}}\right],
\end{equation}
where the propensity score is specified as CTR: $q_{u,i}=\mathbb{P}(o_{u,i}=1)$. 
Notably, $\mathcal{L}_\mathrm{IPS}$ is an unbiased estimator of the ideal learning objective in~\eqref{eq:ideal}, \ie $\mathbb{E}_{O}(\mathcal{L}_\mathrm{IPS})=\mathcal{P}$, when the propensity estimate is accurate.

However, the IPS method can exhibit high variance, particularly in sparse data scenarios like CVR estimation, where propensities are often extremely small. To address this, the doubly robust (DR) estimator \cite{dr} incorporates an error imputation technique. It constructs an imputation model $\hat{\varepsilon}_{u,i}$ to approximate CVR estimation errors in $\mathcal{D}$, and refines the imputation with $\hat{e}_{u,i} = \varepsilon_{u,i} - \hat{\varepsilon}_{u,i}$ in $\mathcal{O}$:
\begin{equation}
\label{eq:dr}
    \mathcal{L}_\mathrm{DR}:=
    \mathbb{E}_{(u, i) \in \mathcal{D}}\left[ \hat{\varepsilon}_{u,i}+\frac{o_{u, i}\hat{e}_{u,i}}{q_{u, i}}\right]
\end{equation}
This formulation ensures unbiasedness as long as either the imputed error $\hat{\varepsilon}_{u,i}$ or the CTR estimate $\hat{o}_{u, i}$ is accurate, hence the term \textit{double robustness}. The IPS and DR estimators offer an unbiased estimation of $\mathcal{P}$ with click data, free from the selection bias introduced by the non-randomness of user clicks. 

\section{Analysis of Entire Space Multitask Model}\label{sec:analyze}
\subsection{Intrinsic Estimation Bias}\label{sec:ieb}

In this section, we delve into the IEB defect with ESMM, where the average CVR estimates exceed the actual values. Previous studies have empirically demonstrated ESMM’s susceptibility to this bias~\cite{mtlips}.
Nonetheless, a formal justification of this defect has not yet been established. We define this problem formally as the IEB problem and establish its presence under mild assumptions in \thmref{thm:ieb}.  

\begin{thm}[Existence of IEB]\label{thm:ieb}
Suppose $O$, $R$, and $C$ are the random variables for click, post-click conversion, and click \& conversion respectively. For a specific user-item pair $(u, i)$, let  $o_{u,i}$,  $r_{u,i}$ , and $c_{u,i}$  denote the actual values; $\hat{o}_{u,i}$,  $\hat{r}_{u,i}$, and  $\hat{c}_{u,i}$  denote the estimated values. The expectation of ESMM’s CVR estimates across all exposures exceeds the true CVR:
\begin{equation}
    \mathrm{Bias}^\mathrm{ESMM}:=
    \mathbb{E}_{\mathcal{D}}\left[\hat{R}\right]-\mathbb{E}_{\mathcal{D}}\left[R\right]>0,
\end{equation}
under the assumption that conversion is more likely to take place for samples within the click space~\cite{treatment}:
\begin{equation*}
    \mathbb{E}_{\mathcal{O}}\left[R\right]>\mathbb{E}_{\mathcal{D}}\left[R\right].
\end{equation*}
\end{thm}

The existence of IEB highlights that sample selection bias cannot be effectively addressed solely by decomposing tasks within the ESMM framework. Instead, it necessitates developing an unbiased estimation approach for the ideal CVR learning objective and directly optimizing it.

\subsection{False Independence Prior}\label{sec:fip}
\begin{figure}
    \centering
    \subfigure[ESMM]{\includegraphics[width=0.26\linewidth,trim=0 0 -3 0,clip]{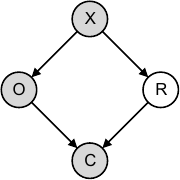}}\hspace{5mm}
    \subfigure[Na\"ive]{\includegraphics[width=0.26\linewidth,trim=0 0 -3 0,clip]{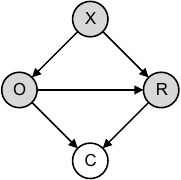}}\hspace{5mm}
    \subfigure[ESCM$^2$]{\includegraphics[width=0.26\linewidth,trim=0 0 -3 0,clip]{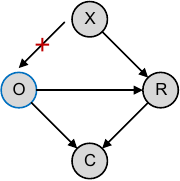}}
    \caption{Causal graphs where X, R, O, C denote the user-item intersection, conversion, click and click \& conversion, respectively. Hollow and shaded nodes indicate latent and observed variables, respectively. The blue circle in (c) represents intervention, blocking the backdoor path X $\rightarrow$ O.}
    \label{fig:causal}
\end{figure}

To estimate CTCVR, ESMM multiplies the outputs of its CTR and CVR arms:
\begin{equation*}
\label{eq:ctcvr1}
    \mathbb{P}(o_{u,i}=1,r_{u,i}=1)=\mathbb{P}(o_{u,i}=1)*\mathbb{P}(r_{u,i}=1\mid o_{u,i}=1),
\end{equation*}
where the CVR estimate is click-dependent, \ie conversion only occurs after the click, establishing a causal link $O \rightarrow R$ in the data generation process. 
However, ESMM's learning objective~\eqaref{eq:esmmLoss} does not explicitly capture this causal dependency, as indicated by the absent arrow $O \rightarrow R$ in \figref{fig:causal} (a).
It poses the risk that ESMM models CVR as $\mathbb{P}(r_{u,i}=1)$ following \figref{fig:causal} (a), as opposed to the expected $\mathbb{P}(r_{u,i}=1\mid o_{u,i}=1)$ in \eqref{eq:ctcvr1}. 
This risk is formulated as false independence prior, as it confuses the targeted $\mathbb{P}(r_{u,i}=1\mid o_{u,i}=1)$ in~\eqaref{eq:ctcvr1} with the unexpected $\mathbb{P}(r_{u,i}=1)$, thereby falsely introducing independent prior in CTCVR estimation.

The na\"ive approach in \eqaref{eq:naive} trains the CVR model within the click space, thereby explicitly incorporating the dependency $O \rightarrow R$ as depicted in \figref{fig:causal} (b)\footnote{This causal graph aligns with Fig. 1 in \cite{gu2021estimating,bareinboim2012controlling}.}.
However, the backdoor path $X \rightarrow O$ introduces sample selection bias \cite{esmm,mtlips}.
From a causality perspective, the key to solving FIP without introducing backdoor path is defining CVR as a causal estimand:
\begin{equation}
\label{eq:cvrcausal}
    \mathbb{P}(r_{u,i}=1\mid \textit{do}(o_{u,i}=1)),
\end{equation}
where "do" represents the do-calculus~\cite{peal_2009}, truncating the backdoor path $X \rightarrow O$ as shown in \figref{fig:causal}.
For clicked samples, the causal estimand \eqref{eq:cvrcausal} aligns with the standard CVR definition, but for unclicked samples, it models the counterfactual problem: \textit{What would be the likelihood of conversion if the user had clicked the item?}.
Based on this formulation, the CTCVR can be redefined as:
\begin{equation*}
\mathbb{P}\left(o_{u,i}=1,r_{u,i}=1\right)=\mathbb{P}\left(o_{u,i}=1\right)*\mathbb{P}\left(r_{u,i}=1\mid do\left(o_{u,i}=1\right)\right),
\end{equation*}
which addresses both FIP and selection bias defects effectively.

\section{Methodology}\label{sec:method}
In this section, we propose ESCM$^2$ to tackle the aforementioned IEB and FIP defects with ESMM.
Section~\ref{sec:crr} describes the implementations and properties of the proposed counterfactual regularizers;
Section~\ref{sec:property} further demonstrates how the proposed regularizers effectively handle the IEB and FIP defects.
Section~\ref{sec:obj} develops ESCM$^2$ by enhancing the ESMM framework with the counterfactual regularizers, detailing the model architecture and learning objectives.

\subsection{Counterfactual Risk Regularizers}\label{sec:crr}

In this section, we contextualize the implementation of two counterfactual risk regularizers: the IPS regularizer and the DR regularizer, elucidating their statistical properties. Given that post-click conversion labels are unavailable for non-clicked samples, a na\"ive approach involves calculating the CVR learning objective based on the estimation errors $\varepsilon_{u,i}$ from clicked samples.  However, this method is prone to sample selection bias, which causes a distribution shift between the training space (click space) and the inference space (exposure space). This shift hinders the CVR estimator's ability to generalize from training to inference, leading to suboptimal performance.

To counteract the distribution shift caused by sample selection bias, the IPS regularizer, as per~\eqaref{eq:ips}, inversely weights each clicked sample (with $o_{u,i}=1$) with propensity score $q_{u,i}$:
\begin{equation}
\begin{aligned}
\label{eq:ipsreg}
\mathcal{R}_\mathrm{IPS}
    &=\mathbb{E}_{(u, i) \in \mathcal{D}}\left[\frac{o_{u, i} \varepsilon_{u,i}}{q_{u, i}}\right]=\frac{1}{|\mathcal{D}|} \sum_{(u, i) \in \mathcal{D}} \frac{o_{u, i} \varepsilon_{u,i}}{\hat{o}_{u, i}},
\end{aligned}
\end{equation}
where the propensity score, typically the actual CTR, is unavailable; hence, the CTR estimate $\hat{o}_{u,i}$ is employed as a proxy \cite{mtlips}.
This re-weighting strategy corrects for the overrepresentation of data that are more prone to be clicked, aligning the training data more closely with the exposure data, thereby addressing the distribution shift between training and inference space. In practice, it offers an approximation of the ideal CVR learning objective—i.e., the expected value of $\varepsilon_{u,i}$ over the dataset $\mathcal{D}$—using data from biased clicked samples. 

The statistical properties of $\mathcal{R}_\mathrm{IPS}$ are encapsulated in~\lemref{thm:ips_bv}. Specifically,  given accurate CTR estimate (\ie $\hat{o}_{u,i}=q_{u,i}$), $\mathcal{R}_\mathrm{IPS}$ is an unbiased estimator (\ie $\mathbb{E}_{O}(\mathcal{R}_\mathrm{IPS})=\mathcal{P}$).
However, $\mathcal{R}_\mathrm{IPS}$ exhibits high variance when $\hat{o}_{u,i}$ values are small, which makes the training process unstable.

\begin{lem}\label{thm:ips_bv}
The bias and variance of $\mathcal{R}_\mathrm{IPS}$ are
\begin{equation*}\label{eq:ipsvar}
    \begin{aligned}
    \operatorname{Bias}_O\left(\mathcal{R}_\mathrm{IPS}\right)&=\frac{1}{|\mathcal{D}|}\left|\sum_{(u, i) \in D} \varepsilon_{u,i}\left(\frac{q_{u,i}}{\hat{o}_{u,i}}-1\right)\right|,\\
    \mathbb{V}_O\left(\mathcal{R}_\mathrm{IPS}\right)&=\frac{1}{|\mathcal{D}|^2} \sum_{(u, i) \in \mathcal{D}} \frac{q_{u,i}\left(1-q_{u,i}\right)}{\hat{o}_{u,i}^2}\left(\varepsilon_{u,i}\right)^2. 
    \end{aligned}
\end{equation*}
\end{lem}

\begin{figure*}
    \centering
    \subfigure[ESCM$^2$-IPS]{\includegraphics[width=0.26\linewidth]{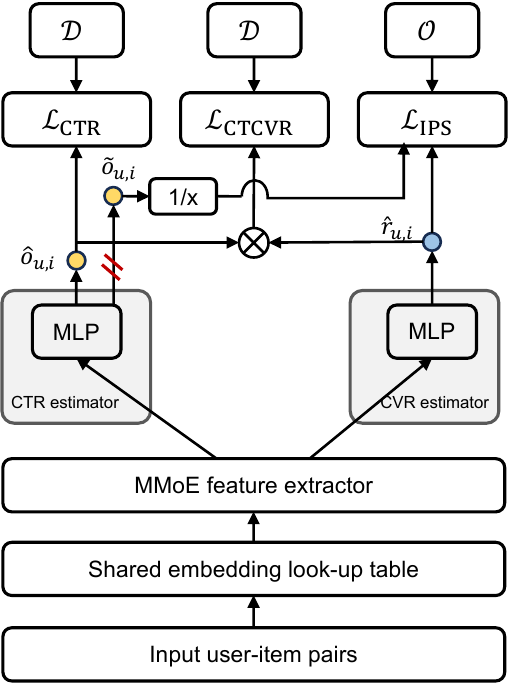}}
    \quad\quad\quad
    \raisebox{-0.0\height}{\rule{0.8pt}{6.2cm}} 
    \quad\quad\quad
    \subfigure[ESCM$^2$-DR]{\includegraphics[width=0.438\linewidth]{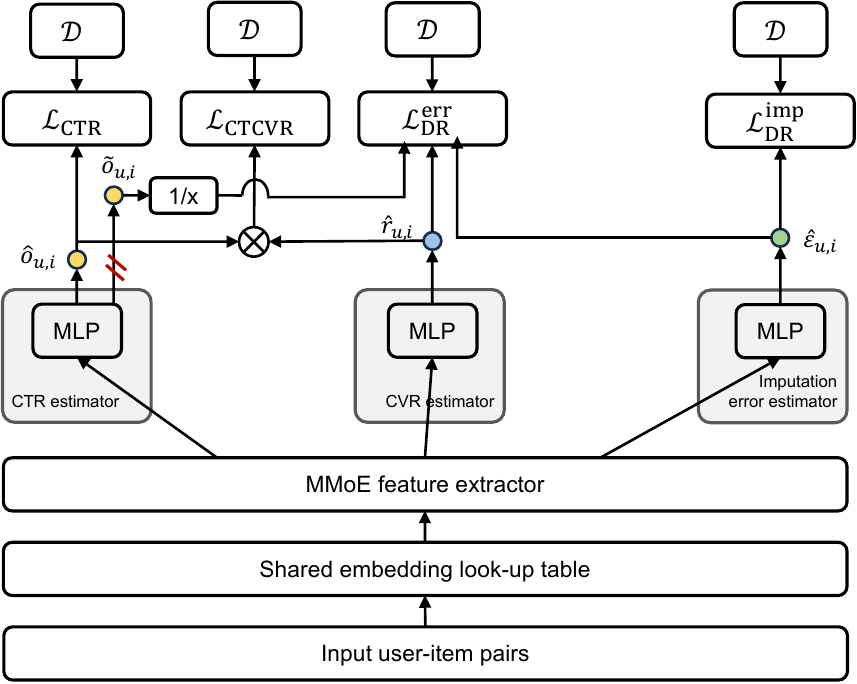}}
    \caption{The core architecture of ESCM$^2$, where the ESCM$^2$-IPS involves two arms for CTR and CVR estimation; the ESCM$^2$-DR involves an additional arm for imputation error estimation. The breaker indicates the path where the  gradients are truncated. $\mathcal{D}$ and $\mathcal{O}$ denote respective exposure and click space.}
    \label{fig:modelStructure}
\end{figure*}

To mitigate the defect with IPS regularizer, the DR regularizer extends $\mathcal{R}_\mathrm{IPS}$ by incorporating an \emph{imputation arm}. This arm aims to accurately impute the CVR estimation error (${\varepsilon}_{u,i}$), and its output, denoted as $\hat{\varepsilon}_{u,i}$, is subsequently corrected by $\hat{e}_{u,i}=\varepsilon_{u,i}-\hat{\varepsilon}_{u,i}$.
The imputation is performed in the exposure space, whereas the correction is executed in the click space where the actual $\varepsilon_{u,i}$ values are available. We implement the DR regularizer as follows:
\begin{equation}
\label{eq:drreg}
\begin{aligned}
\mathcal{R}_\mathrm{DR}^\mathrm{err}
    &=\mathbb{E}_{(u, i) \in \mathcal{D}}\left[\hat{\varepsilon}_{u,i}+\frac{o_{u, i} \hat{e}_{u, i}}{q_{u, i}}\right]\\
    &=\frac{1}{|\mathcal{D}|} \sum_{(u, i) \in \mathcal{D}}\hat{\varepsilon}_{u,i}+\frac{o_{u, i}\hat{e}_{u,i}}{\hat{o}_{u, i}},
\end{aligned}
\end{equation}
where $\hat{e}_{u,i}$ is weighted with the propensity score $q_{u,i}$ to counteract sample selection bias, $\hat{o}_{u,i}$ is a proxy for the propensity score $q_{u,i}$.  
This strategy imputes CVR estimation errors for samples outside the click space and corrects this imputation with $e_{u,i}$. Although $e_{u,i}$ is only available in the click space, the re-weighting corrects for the sample selection bias and offers an equivalent expectation over the exposure space. 

Similar to $\mathcal{R}_\mathrm{IPS}$, $\mathcal{R}_\mathrm{DR}^\mathrm{err}$ is unbiased to $\mathcal{P}$. Moreover, according to~\lemref{thm:dr_bv}, $\mathcal{R}_\mathrm{DR}$ exhibits lower variance than $\mathcal{R}_\mathrm{IPS}$ when $0 < \hat{\varepsilon}_{u,i} < 2\varepsilon_{u,i}$. Besides,  $\mathcal{R}_\mathrm{DR}$ is doubly robust since it ensures unbiasedness if either the propensity estimation or the error imputation is accurate. The accuracy of $\hat{o}_{u,i}$ can be guaranteed by an arbitrary well-trained CTR estimator, and the accuracy of $\hat{\varepsilon}_{u,i}$ can be assured by an auxiliary learning task:
\begin{equation}
\begin{aligned}
\mathcal{R}_\mathrm{DR}^\mathrm{imp}=\mathbb{E}_{(u, i) \in \mathcal{D}}\left[\frac{o_{u,i}\hat{e}^2_{u,i}}{\hat{o}_{u,i}}\right],
\end{aligned}
\end{equation}
and the final learning objective of the DR regularizer is
\begin{equation}\label{eq:dr_final}
\begin{aligned}
\mathcal{R}_\mathrm{DR}&=\mathcal{R}_\mathrm{DR}^\mathrm{err}+\mathcal{R}_\mathrm{DR}^\mathrm{imp}.
\end{aligned}
\end{equation}

\begin{lem}\label{thm:dr_bv}
The bias and variance of $\mathcal{R}_\mathrm{DR}^\mathrm{err}$ are
\begin{equation*}\label{eq:drvar}
    \begin{aligned}
    \operatorname{Bias}\left(\mathcal{R}_\mathrm{DR}^\mathrm{err}\right)&=\frac{1}{|\mathcal{D}|}\left|\sum_{(u, i) \in D}\left(q_{u, i}-\hat{o}_{u, i}\right) \frac{\left(\varepsilon_{u,i}-\hat{\varepsilon}_{u,i}\right)}{\hat{o}_{u, i}}\right|,\\
    \mathbb{V}_O\left(\mathcal{R}_\mathrm{DR}^\mathrm{err}\right)&=\frac{1}{|\mathcal{D}|^2} \sum_{(u, i) \in \mathcal{D}} q_{u, i}\left(1-q_{u, i}\right) \frac{\left(\hat{\varepsilon}_{u,i}-\varepsilon_{u,i}\right)^2}{\hat{o}_{u, i}^2}.
    \end{aligned}
\end{equation*}
\end{lem}

\subsection{Analytical Properties}\label{sec:property}
In this section, we demonstrate that the IPS regularizer ($\mathcal{R}_\mathrm{IPS}$) can effectively handle the IEB and FIP defects with ESMM.
Specifically, \thmref{thm:anti-ieb} establishes that $\mathcal{R}_\mathrm{IPS}$ aligns with the ideal learning objective $\mathcal{P}$ in~\eqaref{eq:ideal}, thereby handling IEB.
Concurrently, \thmref{thm:anti-pip} establishes that $\mathcal{R}_\mathrm{IPS}$ promotes the estimation of the CVR as $\mathbb{P}\left(r_{u, i}=1 \mid do(o_{u, i}=1)\right)$, as specified in \eqaref{eq:cvrcausal}, which explicitly models the causal link from click to conversion, thereby mitigating FIP.
These theoretical results are also applicable to the DR regularizer $\mathcal{R}_\mathrm{DR}$, with detailed discussions provided in Theorems \ref{thm:anti-ieb-dr}-\ref{thm:anti-pip-dr} in the appendix.
\begin{thm}[$\mathcal{R}_\mathrm{IPS}$ handles IEB]
\label{thm:anti-ieb}
Given accurate propensity score estimation, \ie $\hat{o}_{u,i}=q_{u,i}$, we have $\mathcal{R}_\mathrm{IPS} = \mathcal{P}$.
\end{thm}

\begin{thm}[$\mathcal{R}_\mathrm{IPS}$ handles PIP]
\label{thm:anti-pip}
Suppose $\hat{r}^\mathrm{IPS}_{u,i}$ is the CVR estimate that optimizes $\mathcal{R}_\mathrm{IPS}$, $\mathbb{P}\left(r_{u, i}=1 \mid d o\left(o_{u, i}=1\right)\right)$ is the counterfactual conversion rate assuming the user clicked the item. For all samples in the exposure space, $\mathcal{R}_\mathrm{IPS}$ encourages:
\begin{equation*}
    \hat{r}_{u,i}^\mathrm{IPS}\rightarrow\mathbb{P}\left(r_{u, i}=1 \mid d o\left(o_{u, i}=1\right)\right).
\end{equation*}
\end{thm}
\begin{cor}[$\mathcal{R}_\mathrm{IPS}$ v.s. importance sampling]\label{cor:is}
$\mathcal{R}_\mathrm{IPS}$ is typically an importance sampling, which computes the ideal expectation over the exposure space $\mathcal{D}$ using samples from the click space $\mathcal{O}$.
\end{cor}

While \thmref{thm:ips_bv} initially identified the unbiasedness of $\mathcal{R}_\mathrm{IPS}$, \thmref{thm:anti-ieb} presents a stronger unbiasedness, emphasizing a strong alignment with importance sampling principles~\cite{is1}. Building on this, Corollary~\ref{cor:is} posits that $\mathcal{R}_\mathrm{IPS}$ typically operates as an importance sampling procedure. This interpretation suggests that the IPS estimator can be enhanced using advanced importance sampling techniques such as adaptive importance sampling and joint importance sampling~\cite{is1,is2,is3}, to achieve superior statistical properties.

\subsection{Architecture and Learning Objective}\label{sec:obj}

While the proposed counterfactual regularizers effectively approximate the ideal learning objective, they do not  yield a deployable recommendation model. To bridge this gap, we introduce ESCM$^2$, which employs the counterfactual risk regularizers for training recommendation models. The detailed steps are outlined in Algorithm~1 and are explained as follows.

First, we construct a multi-arm estimator $f$ to estimate the CTR, CVR and imputation error (step 1). The architecture of $f$, illustrated in Figure~\ref{fig:modelStructure}, comprises an embedding lookup table, a feature extractor, and task-specific prediction arms. To mitigate data sparsity, the embedding lookup table is shared across different tasks, and $f$ is implemented using a MMoE model~\cite{mmoe}.

Subsequently, we compute the CVR loss $\mathcal{L}_\mathrm{CVR}$ using the counterfactual regularizers (steps~2–6). When the IPS regularizer is employed, $\mathcal{L}_\mathrm{CVR}$ corresponds to $\mathcal{R}_\mathrm{IPS}$ as defined in \eqref{eq:ipsreg}. Crucially, we truncate the gradient of $\mathcal{R}_\mathrm{IPS}$ with respect to $\hat{o}_{u,i}$ (step 2) because the CTR estimate acts only as a coefficient in this context; optimizing it alongside the CVR objective would degrade CTR estimation performance. Alternatively, if the DR regularizer is used, $\mathcal{L}_\mathrm{CVR}$ corresponds to $\mathcal{R}_\mathrm{DR}$ as defined in \eqref{eq:dr_final}. For training stability, we stop the gradient flow from $\hat{\varepsilon}_{u,i}$ when calibrating it with true errors and from $\varepsilon_{u,i}$ when optimizing its estimation.

Finally, we define the learning objective of ESCM$^2$, which comprises three terms (steps 7-8):
\begin{equation}
\label{eq:escm}
\mathcal{L}_\mathrm{ESCM^2}:=\mathcal{L}_\mathrm{CTR}+\lambda_\mathrm{c}\mathcal{L}_\mathrm{CVR}+\lambda_\mathrm{g}\mathcal{L}_\mathrm{CTCVR},
\end{equation}
where $\mathcal{L}_\mathrm{CTR}$ and $\mathcal{L}_\mathrm{CTCVR}$ represent the CTR and CTCVR estimation risks defined in (4); $\mathcal{L}_\mathrm{CVR}$ is the counterfactual CVR risk; $\lambda_\mathrm{c}$ and $\lambda_\mathrm{g}$ are weighting factors. This configuration enables ESCM$^2$ to leverage the ESMM structure to counteract data sparsity while effectively handling the IEB and FIP defects with ESMM using counterfactual regularizers.

\begin{algorithm}[tb]
\caption{The computational procedure for ESCM$^2$.}\label{alg:fws}
\textbf{Input}: 
$(u,i)\in\mathcal{D}$: the user-item pairs in the exposure space; $o_{u,i}$: the click label in the exposure space; $r_{u,i}$: the conversion label in the click space. \\
\textbf{Parameter}: $\lambda_c$: the weight of the counterfactual risk; $\lambda_g$: the weight of the global risk. \\
\textbf{Output}: $\mathcal{L}_\mathrm{ESCM^2}$: the learning objective of ESCM$^2$.
\begin{algorithmic}[1] 
    \State $\hat{o}_{u,i},\hat{r}_{u,i},\hat{\varepsilon}_{u,i} \leftarrow f(u,i)$.
    \State $\tilde{o}_{u,i}\leftarrow \mathrm{StopGradient}(\hat{o}_{u,i})$.
    \If{model is ESCM$^2$-IPS}
    \State Calculate $\mathcal{L}_\mathrm{CVR}$ as $\mathcal{R}_\mathrm{IPS}$ in Eq.\eqref{eq:ipsreg}.
    \ElsIf{ model is ESCM$^2$-DR}
    \State Calculate $\mathcal{L}_\mathrm{CVR}$ as $\mathcal{R}_\mathrm{DR}^\mathrm{err}+\mathcal{R}_\mathrm{DR}^\mathrm{imp}$ in Eq.\eqref{eq:dr_final}.
    \EndIf
    \State Calculate $\mathcal{L}_\mathrm{CTR}$ and $\mathcal{L}_\mathrm{CTCVR}$ in Eq.\eqref{eq:esmmLoss}.
    \State Calculate $\mathcal{L}_\mathrm{ESCM^2}$ in Eq.\eqref{eq:escm}.
\end{algorithmic}
\end{algorithm}

\section{Experiments}\label{sec:exp}
In this section, we conduct experiments to investigate the research questions as follows:
\\\textbf{RQ1:} How does ESCM$^2$ perform compared to the prevalent CVR and CTCVR estimators in offline and online scenarios?
\\\textbf{RQ2:} Does ESMM suffer from the intrinsic estimation bias on CVR estimation? Does ESCM$^2$ effectively reduce the bias?
\\\textbf{RQ3:} Does ESMM suffer from false independence prior in CTCVR estimation? Does ESCM$^2$ mitigate this problem?
\\\textbf{RQ4:} How to tune the weights of learning objectives? 
Is the performance of ESCM$^2$ sensitive to it?
\subsection{Setup}
\subsubsection{Dataset} Experiments are conducted using two datasets, with details summarized in \tabref{tab:datasets}.
    The \textbf{Industry} dataset is constructed using our industrial recommendation logs over 90 days, segmented chronologically into training, validation, and test sets. Negative samples are downsampled in the training phase to maintain an approximate exposure:click:conversion ratio of 100:10:1. The \textbf{Ali-CCP} dataset is incorporated for reproducibility \footnote{\url{https://tianchi.aliyun.com/datalab/dataSet.html?dataId=408}}. Only single-valued categorical fields are used following Xi et al.~\cite{seq2021}, and 10\% of the training set is reserved for validation.

\begin{table}[]\setlength{\tabcolsep}{3pt}
\caption{Dataset description.\label{tab:datasets}}
    \centering
    \begin{tabular}{lllllll}
    \toprule
         Name      &    \# Train  &   \# Valid   &    \# Test &   \# User  & \# Click & \# Conversion \\ \midrule
         Industry    &   61.58M      &   0.39M      & 24.28M &  37.73M 
         & 3.73M & 0.32M
         \\
         Ali-CCP & 33.12M & 3.67M & 37.64M & 0.25M & 1.42M & 7.92K
         \\
         \bottomrule
    \end{tabular}
    
\end{table}

\subsubsection{Baselines}
Given that Multi-Task Learning (MTL) significantly enhances the performance of recommender systems~\cite{mtlips}, single-task CVR estimation approaches~\cite{treatment,dr} are excluded from our baselines to provide a fair comparison. We commence with three prevalent methods that co-train CTR and CVR estimators and share embeddings between them:
\begin{itemize}[leftmargin=*]
    \item \textbf{Naïve}\footnote{\url{https://github.com/PaddlePaddle/PaddleRec/tree/master/models/multitask}\label{mtlmodel}} \cite{mmoe} optimizes the CTR estimator in the exposure space and the CVR estimator in the click space using the biased approach described in \eqaref{eq:naive}.
    \item \textbf{MTL-IMP} \cite{esmm} extends \textbf{Naïve} by including unclicked samples as negative samples to train the CVR estimator.
    \item \textbf{ESMM}\textsuperscript{\ref{mtlmodel}} \cite{esmm} employs a multitask approach to optimize two independent learning objectives for CTCVR and CTR~\cite{esmm}, and implicitly optimizes the CVR estimator.
\end{itemize}

Moreover, we incorporate debiased methods as follows:
\begin{itemize}[leftmargin=*]
    \item \textbf{MTL-EIB} \cite{eib} imputes the CVR estimation error for all samples and corrects its imputation with clicked samples to achieve a theoretically unbiased CVR estimation.
    \item \textbf{MTL-IPS}\footnote{\url{https://github.com/DongHande/AutoDebias/tree/main/baselines}.\label{autodebias}} and \textbf{MTL-DR}~\cite{mtlips} integrates the IPS and DR approach \cite{treatment} into a multitask learning framework, respectively, providing a theoretically unbiased CVR estimator.
\end{itemize}

\subsubsection{Training Protocol}
For all methods in comparison, the multitask estimator is implemented as a standard MMoE model~\cite{mmoe}, beginning with a shared embedding layer. The embedding dimension is uniformly set to 5, with other model settings consistent with standard MMoE.
The learning rate and weight decay are set to 1e$^{-4}$ and 1e$^{-3}$, respectively. Other optimizer settings are consistent with Adam optimizer~\cite{adam}.
Notably, due to the one-epoch saturation phenomenon observed in industrial recommenders \cite{zhang2022towards}, where model performance tends to degrade after more than one epoch of training, each model is trained for a single epoch with batch size 512.
The weighting factors, \(\lambda_\mathrm{g}\) and \(\lambda_\mathrm{c}\), are determined based on the outcomes of a hyperparameter study presented in Section~\ref{sec:param}, with values set to 1 and 0.1. All experiments are conducted on K8S clusters in Ant Group with Intel Xeon Platinum CPUs.

\subsubsection{Evaluation Protocol}
We mainly use the area under the receiver operating characteristic curve (AUC) metric to assess the ranking performance of the models. While AUC is a robust measure for evaluating the average ranking performance across all possible thresholds, it does not provide detailed insights into performance at a specific threshold. Therefore, we supply the KS, recall and F1 metrics at the best thresholds.
Performance is evaluated every 1 thousand iterations on the validation dataset, where the model with the highest AUC is selected for further evaluation on the test dataset. 

\begin{table*}[]
\setlength{\tabcolsep}{3pt}
\caption{Overall performance comparison of CVR and CTCVR estimation (mean±std).}\label{tab:ctcvr}
\footnotesize\centering\renewcommand\arraystretch{1}
\begin{tabular}{l|l|ccccccccc}
\hline
& & \multicolumn{4}{c}{CVR task} & & \multicolumn{4}{c}{CTCVR task} \\
\cline{3-6} \cline{8-11}
Dataset & Model  & Recall      & F1      & KS      & AUC  & & Recall      & F1      & KS      & AUC \\
\hline
\multirow{8}{*}{Industry}
& Naïve    & 0.5789$_{\pm0.0071}$   & 0.3344$_{\pm0.0052}$   & 0.3872$_{\pm0.0043}$   & 0.7515$_{\pm0.0164}$ & & 0.6602$_{\pm 0.0764}$   & 1.0048$_{\pm 0.1751}$  & 0.4631$_{\pm 0.0054}$  & 0.7954$_{\pm 0.0091}$                      \\
& ESMM    & 0.5742$_{\pm0.0055 }$   & 0.6330$_{\pm0.0074}$   & 0.3856$_{\pm0.0051}$   & 0.7547$_{\pm0.0183 }$ &   & \underline{0.6819}$_{\pm 0.0672}$   & 1.1062$_{\pm 0.0527}$  & 0.4827$_{\pm 0.0042}$  & \underline{0.8153}$_{\pm 0.0094}$                      \\
& MTL-EIB    & 0.5121$_{\pm0.0072 }$   & 0.5808$_{\pm0.0048}$   & 0.3371$_{\pm0.0051}$   & 0.7272$_{\pm0.0140 }$   & & 0.5975$_{\pm 0.0590}$   & 0.8458$_{\pm 0.0097}$  & 0.4220$_{\pm 0.0043}$  & 0.7912$_{\pm 0.0108}$                   \\
& MTL-IMP    & \underline{0.5841}$_{\pm0.0092}$   & 0.1272$_{\pm0.0040}$   & 0.3974$_{\pm0.0047}$   & 0.7563$_{\pm0.0114  }$   & & 0.5880$_{\pm 0.0765}$   & \underline{\textbf{1.3393}}$_{\pm 0.0094}$  & 0.4126$_{\pm 0.0047}$  & 0.7752$_{\pm 0.0084}$                 \\
& MTL-IPS    & 0.5651$_{\pm0.0068 }$   & \underline{0.6810}$_{\pm0.0042}$   & 0.3960$_{\pm0.0048}$   & \underline{0.7586}$_{\pm0.0112}$     & & 0.6716$_{\pm 0.0502}$   & 1.0653$_{\pm 0.0081}$  & 0.4840$_{\pm 0.0063}$  & 0.8044$_{\pm 0.0067}$           \\
& MTL-DR     & 0.5137$_{\pm0.0081 }$   & 0.6804$_{\pm0.0042}$   & \underline{0.4016}$_{\pm0.0046}$   & 0.7579$_{\pm0.0135      }$   & & 0.6684$_{\pm 0.0733}$   & 1.1707$_{\pm 0.0061}$  & \underline{0.4844}$_{\pm 0.0044}$  & 0.8106$_{\pm 0.0107}$                \\\cline{3-9}
& \cellcolor{gray!10}ESCM$^2$-IPS   & \cellcolor{gray!10}0.5932$_{\pm0.0094 }$   & \cellcolor{gray!10}\textbf{0.7161}\sig$_{\pm0.0089}$   & \cellcolor{gray!10}\textbf{0.4144}\sig$_{\pm0.0051}$   & \cellcolor{gray!10}\textbf{0.7730}$_{\pm0.0150}$  & \cellcolor{gray!10}& \cellcolor{gray!10}0.6804$_{\pm 0.0771}$   & \cellcolor{gray!10}1.1753$_{\pm 0.0259}$  & \cellcolor{gray!10}0.8198$_{\pm 0.0042}$  & \cellcolor{gray!10}0.7730$_{\pm 0.0062}$   \\
& \cellcolor{gray!10}ESCM$^2$-DR    & \cellcolor{gray!10}\textbf{0.5986}\sig$_{\pm0.0068} $   & \cellcolor{gray!10}0.6884$_{\pm0.0052}$   & \cellcolor{gray!10}0.4119$_{\pm0.0050}$   & \cellcolor{gray!10}0.7679$_{\pm0.0113}$   & \cellcolor{gray!10}& \cellcolor{gray!10}\textbf{0.7013\sig$_{\pm 0.0852}$}   & \cellcolor{gray!10}1.2842$_{\pm 0.0070}$  & \cellcolor{gray!10}\textbf{0.5134\sig$_{\pm 0.0038}$}  & \cellcolor{gray!10}\textbf{0.8265$_{\pm 0.0090}$}   \\
\hline
\multirow{8}{*}{Ali-CCP}
& Naïve      & 0.2854$_{\pm0.0050 }$   & 0.0991$_{\pm0.0053 }$   & 0.1123$_{\pm0.0056}$  & 0.5987$_{\pm0.0139}$  & & 0.2921$_{\pm 0.0044}$   & 0.0978$_{\pm 0.0036}$   & 0.1192$_{\pm 0.0040}$  & 0.6003$_{\pm 0.0133}$  \\
& ESMM        & 0.2968$_{\pm0.0036 }$ & 0.1157$_{\pm0.0084 }$ &  \underline{0.1267}$_{\pm0.0043}$ & 0.6071$_{\pm0.0133}$  & & \underline{0.3027}$_{\pm 0.0032}$   & \underline{0.1139}$_{\pm 0.0050}$   &  0.1292$_{\pm 0.0036}$  & 0.6081$_{\pm 0.0125}$  \\
& MTL-EIB     & 0.2372$_{\pm0.0043 }$ & 0.0825$_{\pm0.0051 }$ & 0.0717$_{\pm0.0057}$ & 0.5603$_{\pm0.0135}$   & & 0.2542$_{\pm 0.0046}$   & 0.0959$_{\pm 0.0039}$   & 0.0697$_{\pm 0.0040}$  & 0.5699$_{\pm 0.0140}$  \\
& MTL-IMP     & 0.2962$_{\pm0.0056 }$ & 0.1135$_{\pm0.0055 }$ & 0.1163$_{\pm0.0043}$ & \underline{0.6114}$_{\pm0.0137}$ & & 0.2973$_{\pm 0.0045}$   & 0.1110$_{\pm 0.0039}$   & 0.1264$_{\pm 0.0038}$  & 0.6087$_{\pm 0.0155}$  \\
& MTL-IPS     & \underline{0.2975}$_{\pm0.0070}$ & 0.0941$_{\pm0.2163 }$ & 0.1177$_{\pm0.0063}$ & 0.6091$_{\pm0.0123}$  & & 0.2991$_{\pm 0.0050}$   & 0.1044$_{\pm 0.0066}$   & 0.1302$_{\pm 0.0028}$  & \underline{0.6138}$_{\pm 0.0155}$  \\
& MTL-DR      & 0.2953$_{\pm0.0178 }$ & \underline{0.1159}$_{\pm0.0084}$ & 0.1255$_{\pm0.0141}$ & 0.6065$_{\pm0.0172}$  & & 0.2980$_{\pm 0.0168}$   & 0.1096$_{\pm 0.0075}$  & \underline{0.1360}$_{\pm 0.0164}$  & 0.6130$_{\pm 0.0192}$  \\\cline{3-9}
& \cellcolor{gray!10}ESCM$^2$-IPS      & \cellcolor{gray!10}0.3061$_{\pm0.0059 }$ & \cellcolor{gray!10}0.1180$_{\pm0.0047 }$ & \cellcolor{gray!10}0.1312$_{\pm0.0060}$ & \cellcolor{gray!10}\textbf{0.6163}$_{\pm0.0151}$  & \cellcolor{gray!10}& \cellcolor{gray!10}\textbf{0.3184\sig$_{\pm 0.0051}$}   & \cellcolor{gray!10}\textbf{0.1207\sig$_{\pm 0.0038}$}   & \cellcolor{gray!10}0.1436$_{\pm 0.0038}$  & \cellcolor{gray!10}0.6189$_{\pm 0.0118}$  \\
& \cellcolor{gray!10}ESCM$^2$-DR       & \cellcolor{gray!10}\textbf{0.3095}\sig$_{\pm0.0054} $ & \cellcolor{gray!10}\textbf{0.1315}\sig$_{\pm0.0053} $ & \cellcolor{gray!10}\textbf{0.1393}\sig$_{\pm0.0042}$ & \cellcolor{gray!10}0.6142$_{\pm0.0133}$  & \cellcolor{gray!10}& \cellcolor{gray!10}0.3117$_{\pm 0.0044}$   & \cellcolor{gray!10}0.1180$_{\pm 0.0040}$   & \cellcolor{gray!10}\textbf{0.1494\sig$_{\pm 0.0038}$}  & \cellcolor{gray!10}\textbf{0.6245$_{\pm 0.0123}$}  \\
\hline

\end{tabular}
\begin{tablenotes}
\item[1] 1. Bold indicates the best performance. The underlined values mark the best performance across all baseline models except ESCM$^2$.
\item[2] 2. “*” marks the significant improvement of the bold metrics relative to underlined metrics, with p-value $<$ 0.01 in the paired samples t-test.
\item[3] 3. Mean and standard deviation are reported over 10 random seeds. F1 scores are reported as percentages to highlight performance differences.
\end{tablenotes}
\end{table*}

\subsection{Overall Performance}
\subsubsection{Performance Evaluation}

We assess ESCM$^2$ against competing models for CVR and CTCVR estimation using offline datasets, as presented in \tabref{tab:ctcvr}. Key observations are summarized as follows:

\begin{itemize}[leftmargin=*]
    \item Debiased baselines generally outperform biased methods. For example, MTL-IPS attains the highest AUC and F1 scores, improving ESMM’s AUC by 0.39\% and KS by 1.04\%. This highlights the potential of integrating unbiased estimators to improve the CVR estimation of ESMM.
    \item ESCM$^2$ significantly enhances performance beyond the best baseline models, attributed to its effective mitigation of the IEB and FIP defects with ESMM, and the benefits of ESMM structure to address data sparsity.
\end{itemize}

In industry, CTCVR is a more commonly used metric for ranking as it encapsulates both clicks and conversions.
Notable findings of CTCVR estimation, as presented in \tabref{tab:ctcvr}, are summarized below:
\begin{itemize}[leftmargin=*]
    \item ESMM demonstrates competitive CTCVR performance, achieving advanced recall and F1 scores on the Ali-CCP dataset and leading recall and AUC scores on the industry dataset. This success is attributed to the inclusion of CTCVR estimation in ESMM’s learning objectives.
    \item ESCM$^2$ outperforms all competitors in CTCVR estimation, albeit with a smaller margin compared to CVR estimation.
    The superiority primarily stems from two factors: (1) incorporating CTCVR learning objective improves the estimation of CTCVR. Secondly, mitigating the IEB and FIP defects through counterfactual regularizers offers more accurate CVR estimate and thereby facilitating CTCVR estimation.
\end{itemize}

\subsection{Online A/B Test}
\begin{table}
\caption{Online A/B test results in 3 scenarios}\label{tab:online1}
\setlength{\tabcolsep}{5pt}\footnotesize\centering\renewcommand\arraystretch{1}
\begin{tabular}{lllllll}
\hline
Scenario       & \# UV & \# PV & \# Order & \# Premium & CVR &CTCVR      \\
\hline
1 & 2.2M & 3.1M & +2.84\% & +10.85\% & +5.64\% & +3.92\% \\
2 & 3.4M & 4.9M & +4.26\% & +3.88\% & +0.43\% & +1.75\% \\
3 & 125K & 136K & +40.55\% & +12.69\% &- &-\\
\hline
\end{tabular}
\end{table}

\begin{table}\renewcommand{\arraystretch}{1}
\caption{Comprehensive Online A/B test results in Scenario 1}\label{tab:online}
\footnotesize\setlength{\tabcolsep}{2pt}\renewcommand\arraystretch{1}
\centering
\begin{tabular}{lllllll}
\hline
Metrics       & Day 1     & Day 2    & Day 3    & Day 4     & Day 5    & Day 6      \\
\hline
\# Order    & $-$9.76\%           & $-$1.85\% & $-$1.43\%  & \textbf{$+$9.07\%}  & \textbf{$+$0.73\%}           & \textbf{$+$6.26\%} \\
\# Premium  & \textbf{$+$64.53\%}  & \textbf{$+$37.47\%} & \textbf{$+$22.09\%} & $-$12.49\%           & \textbf{$+$4.26\%}  & \textbf{$+$11.10\%} \\
UV$-$CVR      & \textbf{$+$7.25\%}   & $-$1.66\%           & \textbf{$+$9.39\%}  & \textbf{$+$8.58\%}   & \textbf{$+$2.51\%}  & \textbf{$+$8.62\%}  \\
UV$-$CTCVR    & \textbf{$+$0.20\%}   & $-$3.50\%          & \textbf{$+$2.50\%}  & \textbf{$+$9.48\%}   & \textbf{$+$2.75\%}  & \textbf{$+$6.64\%}  \\
\hline
\end{tabular}
\end{table}

To further demonstrate the advantage of ESCM$^2$ over ESMM, online experiments are conducted on the industrial recommendation systems in Alipay.
We first implement ESMM and ESCM$^2$\footnote{ESCM$^2$ is implemented with the IPS regularizer for advantageous training efficiency.} using our C$++$ based machine learning engine and open data processing service. 
Then, unique visitors (UV) are assigned to either ESMM or ESCM$^2$, and performance is compared using four metrics: UV-CVR, UV-CTCVR, order quantity (\# Order) and total premium (\# Premium). 
Experiments across three large-scale scenarios, summarized in \tabref{tab:online1}, yielded the following results:

\begin{itemize}[leftmargin=*]
    \item \textbf{Scenario 1.} This scenario is the insurance recommendation from Alipay. Over six days with 3.1M page views (PVs) and 2.2M UVs, ESCM$^2$ increased the total premium by 10.85\%, order quantity by 2.84\% and UV-CVR by 5.64\%. Daily comparisons are performed in \tabref{tab:online}, where ESCM$^2$ outperforms ESMM in most metrics. 
    
    \item \textbf{Scenario 2.} This scenario is a renovation of the scenario above, where ESCM$^2$ boosted the total premium by 3.88\%, order quantity by 4.26\%, UV-CVR by 0.43\%, and UV-CTCVR by 1.75\%.
    
    \item \textbf{Scenario 3.} The third scenario is the Wufu campaign in Alipay, where ESCM$^2$ achieved a 40.55\% increase in order quantity and a 12.69\% rise in premium.
\end{itemize}

\subsection{Additional Study on Intrinsic Estimation Bias}
In this section, we examine the IEB defect in ESMM and assess the effectiveness of ESCM$^2$ in addressing it. We compare the average CVR labels ($\bar{r}$) with the model estimates ($\tilde{r}$) as shown in \tabref{tab:cvrbias}. Since conversion labels are only available within the click space, $\bar{r}$ serves as an upper bound for the actual average CVR across the entire exposure space. Thus, any deviation from $\bar{r}$ provides a lower-bound estimate of the CVR estimation bias.

\tabref{tab:cvrbias} indicates that ESMM consistently overestimates CVR values, validating the theoretical result in \thmref{thm:ieb} and confirming the presence of IEB.
Conversely, ESCM$^2$ significantly mitigates CVR estimation bias. On the industry dataset, ESCM$^2$-IPS achieves a bias reduction of 48.95\% and 28.55\% in the training and test sets, respectively, while ESCM$^2$-DR achieves reductions of 58.58\% and 36.81\%.Similar improvements are observed on the Ali-CCP dataset, with ESCM$^2$-IPS and ESCM$^2$-DR reducing bias by comparable percentages, highlighting the general effectiveness of ESCM$^2$ in addressing IEB. This efficacy is attributed to the counterfactual regularizers in ESCM$^2$, which ensure unbiased CVR estimates, as demonstrated in \thmref{thm:anti-ieb}

\begin{table*}[]\renewcommand\arraystretch{1.3}\footnotesize\centering
\caption{Overview of the IEB defect with ESMM and the efficacy of ESCM$^2$ to mitigate IEB.}
\label{tab:cvrbias}
\setlength{\tabcolsep}{3pt}
\begin{tabular}{ccc|cc|ccc|ccc}
\hline
\multirow{2}*{Dataset} & \multirow{2}*{Subset} & \multirow{2}*{$\bar{r}$} & \multicolumn{2}{c}{ESMM} & \multicolumn{3}{c}{ESCM$^2$-IPS} & \multicolumn{3}{c}{ESCM$^2$-DR} \\\cline{4-11}
&&& $\tilde{r}$ & $|\tilde{r}-\bar{r}|$ &  $\tilde{r}$ & $|\tilde{r}-\bar{r}|$ & $\Delta (\%)$ &  $\tilde{r}$ & $|\tilde{r}-\bar{r}|$ & $\Delta (\%)$ \\
\hline
Ali-CCP  & Train   & 0.0055      & 0.0101$_{\pm 0.0011}$  &  0.0046$_{\pm 0.0011}$   &  0.0059$_{\pm 0.0005}$   &0.0004$_{\pm 0.0005}$& \textcolor{abl}{91.30\%$\downarrow$} &  0.0076$_{\pm 0.0018}$ & 0.0021$_{\pm 0.0018}$  & \textcolor{abl}{54.34\%$\downarrow$} \\
Ali-CCP  & Test    & 0.0056      & 0.0113$_{\pm 0.0008}$  &  0.0057$_{\pm 0.0008}$   &  0.0060$_{\pm 0.0009}$   &0.0004$_{\pm 0.0009}$& \textcolor{abl}{92.98\%$\downarrow$} &  0.0059$_{\pm 0.0009}$ & 0.0003$_{\pm 0.0009}$  & \textcolor{abl}{94.73\%$\downarrow$} \\
Industry & Train   & 0.0953      & 0.1588$_{\pm 0.0107}$  &  0.0635$_{\pm 0.0107}$   &  0.1277$_{\pm 0.0053}$   &0.0324$_{\pm 0.0053}$& \textcolor{abl}{48.97\%$\downarrow$} &  0.1216$_{\pm 0.0093}$ & 0.0263$_{\pm 0.0093}$  & \textcolor{abl}{58.58\%$\downarrow$} \\
Industry & Test    & 0.0407      & 0.1643$_{\pm 0.0095}$  &  0.1236$_{\pm 0.0095}$   &  0.1290$_{\pm 0.0092}$   &0.0883$_{\pm 0.0092}$& \textcolor{abl}{28.55\%$\downarrow$} &  0.1188$_{\pm 0.0022}$ & 0.0781$_{\pm 0.0022}$  & \textcolor{abl}{36.81\%$\downarrow$} \\
\hline
\end{tabular}
\begin{tablenotes}
\item[1] 1. $\bar{r}$ indicates the average of the CVR label in the click space, calculated as the quotient of the number of samples: $|\mathcal{O}|/|\mathcal{D}|$.
\item[2] 2. $\tilde{r}$ indicates the average of the model's CVR estimates for all samples in the dataset. Its bias from the mean CVR ground truth is denoted as $|\tilde{r}-\bar{r}|$.
\item[3] 3. $\Delta$ denotes the relative reduction in the model's bias compared to the ESMM.
\end{tablenotes}
\end{table*}

\subsection{Additional Study on False Independence Prior}

In this analysis, we investigate the FIP defect in ESMM and assess how ESCM$^2$ addresses it. FIP arises since the model fails to capture the causal link between clicks and conversions, as illustrated by the missing $O \rightarrow R$ in \figref{fig:causal} (a).
To demonstrate FIP, we measure the causation strength from clicks to conversions using propensity score matching, which accounts for the confounding effect of $X$ in \figref{fig:causal}. In this analysis, CVR and CTR estimates are treated as outcomes and propensities, respectively, to study the causal link strength. Following \cite{psm}, we divide samples into clicked and unclicked groups based on click labels and pair clicked samples with unclicked ones that have the most similar propensity scores. The causal link strength $O \rightarrow R$ is estimated using the causal risk ratio (CRR)~\cite{whatif}, where a CRR close to 1 indicates weak causation. Thus, the causation strength is quantified as $|\mathrm{CRR} - 1|$.

\begin{figure}
    \centering
    \includegraphics[width=\linewidth]{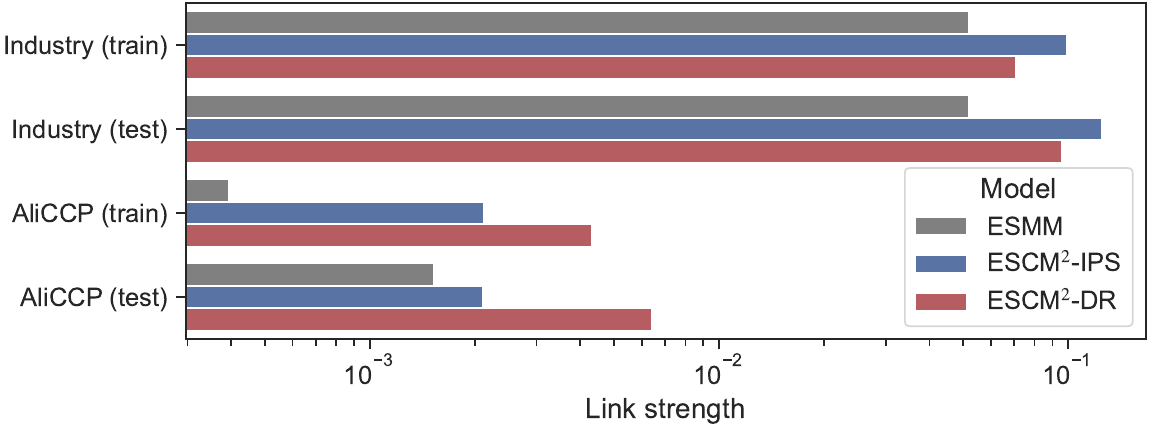}
    \caption{Comparative study of the causal link strength $O \rightarrow R$ with and without counterfactual regularizers.}
    \label{fig:causality}
\end{figure}

\figref{fig:causality} shows that ESMM exhibits minimal causation strength, approximately 0.05 and 0.001 on the industry and Ali-CCP datasets, respectively, confirming the presence of the FIP defect. In contrast, ESCM$^2$ significantly increases causation strength, with ESCM$^2$-IPS achieving over 0.12 and 0.002 on the respective datasets. This improvement is attributed to ESCM$^2$'s counterfactual regularizers, which estimate CVR as $\mathbb{P}\left(r_{u,i}=1 \mid do(o_{u,i}=1)\right)$ according to \thmref{thm:anti-pip}, , explicitly accounting for the causal effect of clicks on conversions and mitigating the FIP defect.

\subsection{Hyper-parameter Tuning and Ablation Study}\label{sec:param}
\begin{figure}
    \centering
\includegraphics[width=\linewidth,trim=20 10 20 10]{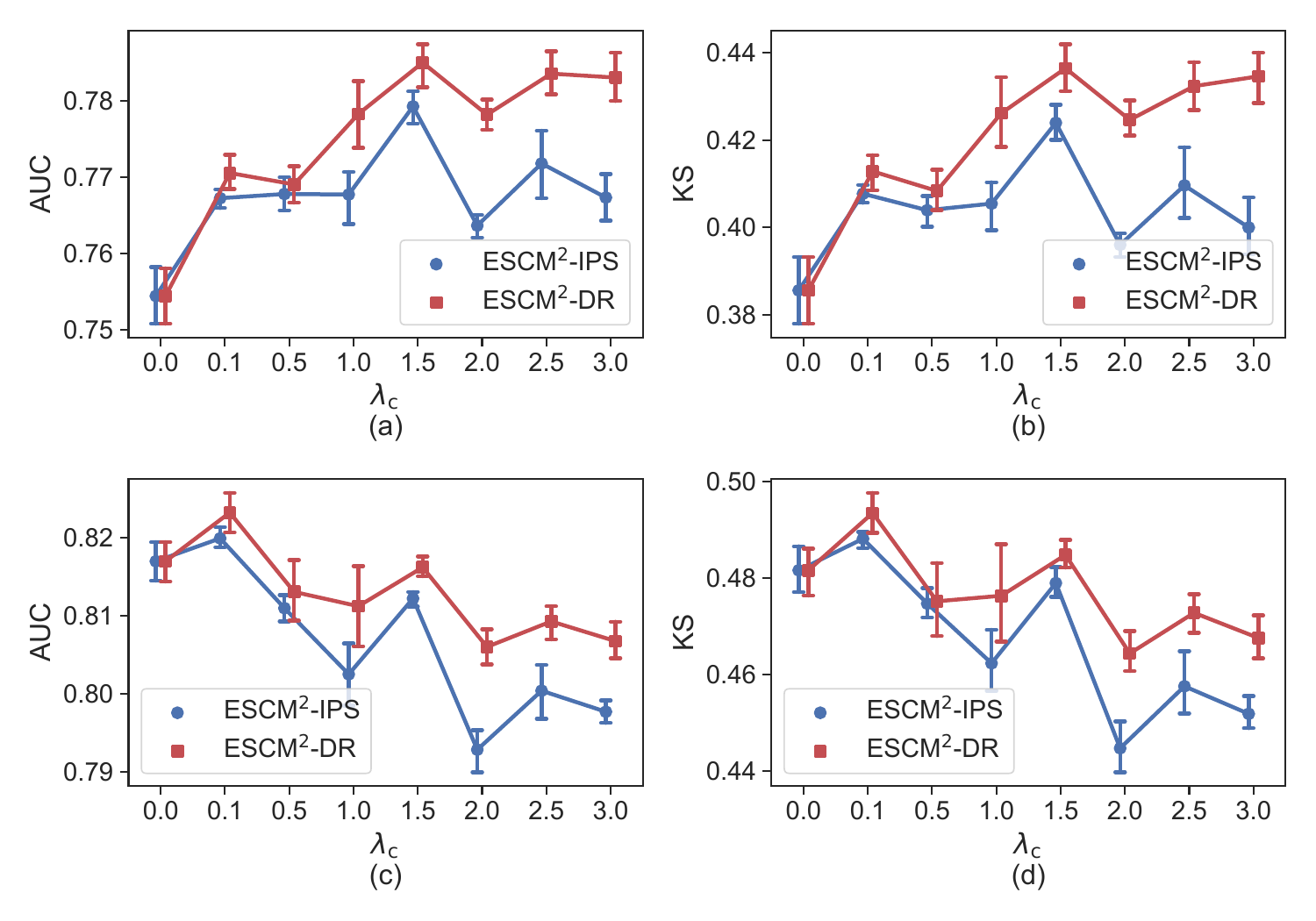}
    \caption{Performance of CVR (a-b) and CTCVR estimation (c-d) with varying counterfactual risk weight $\lambda_\mathrm{c}$.}
    \label{fig:lambda_c}
\end{figure}

\begin{figure}
    \centering
\includegraphics[width=\linewidth,trim=20 10 20 10]{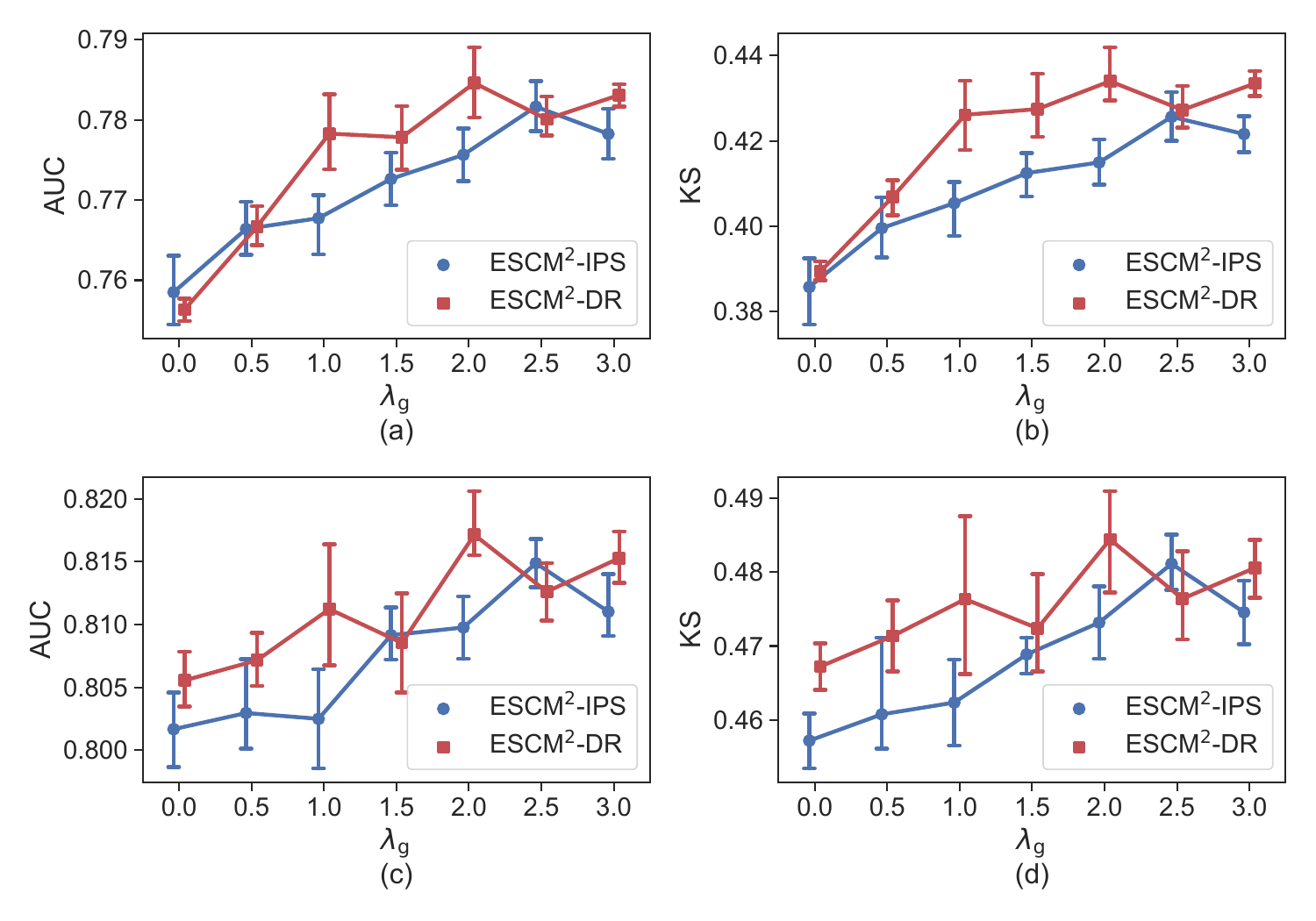}
    \caption{Performance of CVR (a-b) and CTCVR estimation (c-d) with varying CTCVR risk weight $\lambda_\mathrm{g}$.}
    \label{fig:lambda_g}
\end{figure}

Two crucial hyperparameters of ESCM$^2$ are the weighting factors (\ie $\lambda_\mathrm{c}$ and $\lambda_\mathrm{g}$) in the learning objective \eqaref{eq:escm}.
In this section, they are tuned within the range [0, 3] to investigate the impact of causal regularization and global risk minimization on the performance of CVR and CTCVR estimation.
\begin{itemize}[leftmargin=*]
    \item The weighting factor $\lambda_\mathrm{c}$ is investigated in \figref{fig:lambda_c}. 
    Evidently, increasing $\lambda_\mathrm{c}$ consistently benefits CVR estimation, which showcases the effectiveness of causal regularization.
    The AUC of ESCM$^2$-DR, for instance, grows from 0.755 at $\lambda_\mathrm{c}=0$ where the causal regularization is not applied, to around 0.785 at $\lambda_\mathrm{c}=1.5$.
    Additionally, causal regularization also benefits CTCVR estimates.
    For example, the AUC of ESCM$^2$-IPS boosts from 0.817 at $\lambda_\mathrm{c}=0$ to 0.821 at $\lambda_\mathrm{c}=0.1$.
    Nevertheless, the overemphasis on CVR risk has a negative impact on CTCVR estimation.
    For example, a drop in AUC by 0.012 is observed for ESCM$^2$-IPS from $\lambda_\mathrm{c}=0.1$ to $\lambda_\mathrm{c}=3.0$.
    This phenomenon is attributed to the seesaw effect in multitask learning~\cite{huang2022modeling}, \ie overemphasis on the CVR risk misleads the optimizer to ignore the CTR risk, reducing CTR and CTCVR estimation performance.
    Therefore, we suggest tuning $\lambda_\mathrm{c}$ within the range [0, 0.1].
    \item The weighting factor $\lambda_\mathrm{g}$ is studied in \figref{fig:lambda_g}. 
    Overall, increasing $\lambda_\mathrm{g}$ within the range [0, 3] is beneficial for both CTR and CTCVR estimation.
    In the CVR estimation task, for instance, increasing $\lambda_\mathrm{g}$ from 0 to 2.5, the KS of ESCM$^2$-DR climbs from 0.385 to about 0.434; the AUC of ESCM$^2$-IPS grows by 0.023, significantly.
    These observations verify the effectiveness of entire space multitask modelling paradigm, exploiting the sequential user behavior track as per~\figref{fig:yangcong}.
\end{itemize}

\section{Related work}\label{sec:rw}

Post-click conversion rate (CVR) estimation is a crucial task in recommendation, which aims to predict the likelihood of a user completing a transaction after clicking~\cite{esmm,chan2023capturing,chen2020esam}. Accurate CVR estimation not only helps to drive transactions, but also improves traffic resource allocation, resulting in increased revenue. However, CVR estimation faces significant challenges, namely sample selection bias and data sparsity~\cite{mtlips,esmm}. 
To address these challenges, existing methods generally fall into two paradigms: the entire space multitask paradigm and the causal recommendation paradigm. 

\subsection{Entire Space Multitask Paradigm}

The entire space multitask paradigm, pioneered with ESMM~\cite{esmm}, innovatively avoids training CVR estimator directly. Instead, it employs a multitask approach to optimize two independent learning objectives for CTCVR and CTR~\cite{esmm}. This strategy leverages all exposure samples for training both CTCVR and CTR objectives, thereby alleviating data sparsity and enhancing practical performance.
On the basis of ESMM, a line of work initiated in~\cite{esm2} advocates for including additional conversion-related actions, such as adding to favorites and cart. These actions enrich the user behavior track and introduce auxiliary tasks, providing more granular supervision labels to further mitigate data sparsity~\cite{yang2022multi,shen2022mbn,jin2022multi}. Subsequently, graph models have been explored as alternatives to the Markov chain approach for capturing increasingly complex interactions among user behaviors~\cite{gmcm,hm3}. In another line of work, various plug-ins, such as language models~\cite{wu2022multi} and the delayed feedback calibrator~\cite{zhao2023entire}, have been integrated into the training paradigm to improve recommendation quality.

This paradigm is particularly prevalent in industrial recommendation systems where accurate CVR estimation is crucial. The data sparsity issue is effectively mitigated by this paradigm through the incorporation of auxiliary tasks with abundant data; however, sample selection bias persists due to the absence of unbiasedness guarantee~\cite{mtlips} and overly simplistic dependency assumptions. These concerns are formulated as two defects with ESMM in this study, namely the intrinsic estimation bias and the false independence prior, which renders ESMM's CVR estimation biased and leaves room for further improvement.

\subsection{Causal Recommendation Paradigm}

The causal recommendation paradigm, initiated with \cite{treatment}, estimates the ideal learning objective by adjusting the biased dataset using propensity scores. 
On this basis, a line of work focuses on enhancing the estimation of propensity scores, which is crucial for ensuring the unbiasedness of the adjusted learning objective~\cite{li2022stabilized}. Early methods estimated the propensity score based on heuristic item popularity~\cite{saito2020unbiased, joachims2017unbiased}, and later progressed to parametric models like logistic regression~\cite{mtlips, dual}.
Subsequent advancements have incorporated various learning techniques, such as feature selection~\cite{shortreed2017outcome}, joint optimization~\cite{mtlips,zhanguser}, alternative training~\cite{zhu2020unbiased} and kernel balancing~\cite{li2024kernel}, for enhanced identifiability and estimation quality.
Another line of works advocates for more complex causal adjustment approaches to reduce estimation variance~\cite{mrdr}, improve training stability~\cite{li2022stabilized}, resist noisy labels~\cite{li2024debiased} and model mis-specification~\cite{lirelaxing}. Some recent works innovatively incorporate a small subset of unbiased data during training~\cite{chen2021autodebias, li2023balancing, li2024removing}, which effectively addresses missing confounders while minimizing additional data collection efforts.

While this paradigm is widely acknowledged in academic research, its application in industrial recommendation scenarios remains limited. Although the sample selection bias is effectively addressed through causal adjustment with solid theoretical guarantees; the training only involves treated (\ie clicked) samples which are sparse in real-world applications. As a result, the issue of data sparsity is ignored by this paradigm, which severely compromises the performance of CVR estimators in industrial practice.

In conclusion, the entire space multitask paradigm and the causal recommendation excel handling data sparsity and sample selection bias, respectively, but fall short of tackling both challenges concurrently. We innovatively synthesize the two paradigms and construct ESCM$^2$, which incorporates a counterfactual risk regularizer within the ESMM framework to regularize CVR estimation. It leverages the advantage of ESMM for mitigating data sparsity, while addressing sample selection bias by estimating unbiased CVR through counterfactual regularizers.

\section{Conclusion}\label{sec:conclusion}
The entire space multitask model is prevalent in industrial CVR estimation, yet it often overlooks IEB and PIP issues. This study handles these defects by integrating a counterfactual risk regularizer within the ESMM framework.
This strategy retains ESMM’s capability to alleviate data sparsity while addressing sample selection bias through counterfactual regularizers, thereby improving the accuracy of CVR estimation.
Real-world experiments demonstrate that the proposed method effectively alleviates both IEB and FIP defects, improving CVR estimation performance.

\textbf{Limitation \& future works.} In this work, we focus on the sequential user behavior track illustrated in~\figref{fig:yangcong}. However, in industrial scenarios, there are diverse user behaviors whose dependencies excess the expressive capabilities of Markov chains.
Although some studies~\cite{esm2,gmcm} extend ESMM to describe such complex dependencies via decomposition, they also inevitably suffer from both IEB and FIP defects.
Leveraging the counterfactual regularization techniques in ESCM$^2$, these methods could be further enhanced to effectively address both IEB and FIP defects.

\textbf{Acknowledgement.}
This work was supported in
part by National Natural Science Foundation of China (623B2002, 62073288, 12075212, 12105246). 
We are also pleased to thank the Ant Group for providing their expertise in the data preparation, offline evaluation and online A/B test phases.

\small
\bibliographystyle{plain}
\bibliography{sample-base}

\clearpage
\appendix
\setcounter{page}{1}
\setcounter{thm}{0}
\subsection{Theoretical justification}

\begin{thm}[Existence of IEB]
Suppose $O$, $R$, and $C$ are the random variables for click, post-click conversion, and click \& conversion respectively. For a specific user-item pair $(u, i)$, let  $o_{u,i}$,  $r_{u,i}$ , and $c_{u,i}$  denote the actual values; $\hat{o}_{u,i}$,  $\hat{r}_{u,i}$, and  $\hat{c}_{u,i}$  denote the estimated values. The expectation of ESMM’s CVR estimates across all exposures exceeds the true CVR:
\begin{equation}
    \mathrm{Bias}^\mathrm{ESMM}:=
    \mathbb{E}_{\mathcal{D}}\left[\hat{R}\right]-\mathbb{E}_{\mathcal{D}}\left[R\right]>0,
\end{equation}
under the assumption that conversion is more likely to take place for samples within the click space~\cite{treatment}:
\begin{equation}\label{eq:esmmasapp}
    \mathbb{E}_{\mathcal{O}}\left[R\right]>\mathbb{E}_{\mathcal{D}}\left[R\right].
\end{equation}
\end{thm}
\begin{proof}
According to the learning objectives in \eqaref{eq:esmmLoss}, one properly trained ESMM guarantees:
\begin{equation}
\begin{aligned}
    \mathbb{E}_{\mathcal{D}}\left[O-\hat{O}\right]&=\int \left(o_{u,i}-\hat{o}_{u,i}\right) \dd(u,i)=0\\
    \mathbb{E}_{\mathcal{D}}\left[C-\hat{C}\right]&=\int \left(c_{u,i}-\hat{c}_{u,i}\right) \dd(u,i)=0.
\end{aligned}
\label{eq:esmm2}
\end{equation}

Recall that $\mathbb{E}_{\mathcal{D}}\left[R\right]$ and $\mathbb{E}_{\mathcal{D}}\left[\hat{R}\right]$ are the expected values of the true CVR and its estimates, respectively. The bias of CVR estimate can be expressed as
\begin{subequations}
\begin{align}
\mathrm{Bias}^\mathrm{ESMM}&=\mathbb{E}_{\mathcal{D}}\left[\hat{R}\right]-\mathbb{E}_{\mathcal{D}}\left[R\right]\notag\\
&>\mathbb{E}_{\mathcal{D}}\left[\hat{R}\right]-\mathbb{E}_{\mathcal{O}}\left[R\right] \label{eq:bias1}\\
&=\mathbb{E}_{\mathcal{D}}\left[\hat{R}\right]-\frac{\mathbb{E}_{\mathcal{D}}\left[C\right]}{\mathbb{E}_{\mathcal{D}}\left[O\right]}\label{eq:bias2}\\
&=\mathbb{E}_{\mathcal{D}}\left[\frac{\hat{C}}{\hat{O}}\right]-\frac{\mathbb{E}_{\mathcal{D}}\left[C\right]}{\mathbb{E}_{\mathcal{D}}\left[O\right]}\label{eq:bias3},
\end{align}
\end{subequations}
where 
\eqaref{eq:bias1} exploits the assumption in \eqaref{eq:esmmasapp}; 
\eqaref{eq:bias2} decomposes the expectation of actual CVR over $\mathcal{O}$ using the user behavior track in \figref{fig:yangcong}, 
\eqaref{eq:bias3} follows the decomposition in \eqaref{eq:esmmDecom}.

In exposure space, denote the joint probability of $\hat{C}=\hat{c}$ and $\hat{O}=\hat{o}$ as $\mathbb{P}(\hat{c}, \hat{o})$. The first term in \eqaref{eq:bias3} can be rewritten as
\begin{subequations}
\begin{align}
\mathbb{E}_{\mathcal{D}}\left[\frac{\hat{C}}{\hat{O}}\right] 
&= \int \frac{\hat{c}}{\hat{o}}\mathbb{P}(\hat{c}, \hat{o})d(\hat{c},\hat{o}) \label{eq:jensen1}\\
&= \int \frac{\hat{c}}{\hat{o}}\mathbb{P}(\hat{c})\mathbb{P}(\hat{o})d(\hat{c},\hat{o}) \label{eq:jensen2}\\
&= \int\hat{c}\mathbb{P}(\hat{c})d\hat{c}\int\frac{1}{\hat{o}}\mathbb{P}(\hat{o})d\hat{o} \label{eq:jensen3}\\
&= \mathbb{E}_{\mathcal{D}}\left[\hat{C}\right]\mathbb{E}_{\mathcal{D}}\left[\frac{1}{\hat{O}}\right] \label{eq:jensen4}\\
&\geq \frac{\mathbb{E}_{\mathcal{D}}\left[\hat{C}\right]}{\mathbb{E}_{\mathcal{D}}\left[\hat{O}\right]} \label{eq:jensen5}\\
&= \frac{\mathbb{E}_{\mathcal{D}}\left[C\right]}{\mathbb{E}_{\mathcal{D}}\left[O\right]}. \label{eq:jensen6}
\end{align}
\end{subequations}
Below is step-by-step derivation:
\begin{itemize}
	\item \eqaref{eq:jensen1} expands the expectation of $\hat{C}/\hat{O}$;
	\item \eqaref{eq:jensen2} holds under a stringent assumption $\hat{O}\upmodels\hat{C}$;
    \item \eqaref{eq:jensen3} decomposes the integrals and prompts \eqaref{eq:jensen4};
    \item \eqaref{eq:jensen5} exploits the Jensen inequality $\mathbb{E}\left[f(X)\right]\geq f\left(\mathbb{E}\left[X\right]\right)$ for convex function $f(X)=1/X$; where the equality holds only when the variance of $X$ is zero;
    \item \eqaref{eq:jensen6} is ensured by a properly trained ESMM in~\eqaref{eq:esmm2}.
 \end{itemize}
 
As a result, ESMM's CVR estimates are always higher than the actual values, \ie $\mathrm{Bias}^\mathrm{ESMM}>0$. The proof is completed.
\end{proof}

\setcounter{thm}{1}
\begin{lem}
The bias and variance of $\mathcal{R}_\mathrm{IPS}$ are
\begin{equation*}
    \begin{aligned}
    \operatorname{Bias}\left(\mathcal{R}_\mathrm{IPS}\right)&=\frac{1}{|\mathcal{D}|}\left|\sum_{(u, i) \in D} \varepsilon_{u,i}\left(\frac{q_{u,i}}{\hat{o}_{u,i}}-1\right)\right|,\\
    \mathbb{V}_O\left(\mathcal{R}_\mathrm{IPS}\right)&=\frac{1}{|\mathcal{D}|^2} \sum_{(u, i) \in \mathcal{D}} \frac{q_{u,i}\left(1-q_{u,i}\right)}{\hat{o}_{u,i}^2}\left(\varepsilon_{u,i}\right)^2. 
    \end{aligned}
\end{equation*}
\end{lem}
\begin{proof}
Recalling that $\mathcal{P}$ is the ideal CVR estimation error in~\eqref{eq:ideal}, $q_{u,i}$ is the propensity score, $\hat{o}_{u,i}$ is the CTR estimate, $\varepsilon_{u,i}$ is the CVR estimation error in~\eqref{eq:cvrerrormeasure}. We have
\begin{equation*}
    \begin{aligned}
    \operatorname{Bias}\left[\mathcal{R}_\mathrm{IPS}\right] 
    &=\left|\mathbb{E}_O\left[\mathcal{R}_\mathrm{IPS}\right]-\mathcal{P}\right| \\ 
    &=\left| \frac{1}{|\mathcal{D}|} \sum_{(u, i) \in \mathcal{D}} \mathbb{E}_O\left[\frac{o_{u, i}\varepsilon_{u,i}}{\hat{o}_{u,i}}\right]-\varepsilon_{u,i}\right|\\ 
    &=\left|\frac{1}{|\mathcal{D}|} \sum_{(u, i) \in \mathcal{D}}\left[\frac{q_{u,i}\varepsilon_{u,i}}{\hat{o}_{u,i}}-\varepsilon_{u,i}\right]\right| \\ 
    &=\frac{1}{|\mathcal{D}|}\left|\sum_{(u, i) \in D} \varepsilon_{u,i}\left(\frac{q_{u,i}}{\hat{o}_{u,i}}-1\right)\right|,
    \end{aligned}
\end{equation*}

\begin{equation*}
    \begin{aligned}
     \mathbb{V}_O\left[\mathcal{R}_\mathrm{IPS}\right] 
     &=\frac{1}{|\mathcal{D}|^2} \sum_{(u, i) \in \mathcal{D}} \mathbb{V}_O\left[\frac{o_{u, i}\varepsilon_{u,i}}{\hat{o}_{u,i}}\right] \\ &=\frac{1}{|\mathcal{D}|^2} \sum_{(u, i) \in \mathcal{D}} \mathbb{V}_O\left[o_{u, i}\right] \cdot\left(\frac{\varepsilon_{u,i}}{\hat{o}_{u,i}}\right)^2\\ 
     &=\frac{1}{|\mathcal{D}|^2} \sum_{(u, i) \in \mathcal{D}} \frac{q_{u,i}\left(1-q_{u,i}\right)}{\hat{o}_{u,i}^2}\left(\varepsilon_{u,i}\right)^2. 
    \end{aligned}
\end{equation*}
\end{proof}

\begin{lem}
The bias and variance of $\mathcal{R}_\mathrm{DR}^\mathrm{err}$ are
\begin{equation*}
    \begin{aligned}
    \operatorname{Bias}\left(\mathcal{R}_\mathrm{DR}^\mathrm{err}\right)&=\frac{1}{|\mathcal{D}|}\left|\sum_{(u, i) \in D}\left(q_{u, i}-\hat{o}_{u, i}\right) \frac{\left(\varepsilon_{u,i}-\hat{\varepsilon}_{u,i}\right)}{\hat{o}_{u, i}}\right|,\\
    \mathbb{V}_O\left(\mathcal{R}_\mathrm{DR}^\mathrm{err}\right)&=\frac{1}{|\mathcal{D}|^2} \sum_{(u, i) \in \mathcal{D}} q_{u, i}\left(1-q_{u, i}\right) \frac{\left(\hat{\varepsilon}_{u,i}-\varepsilon_{u,i}\right)^2}{\hat{o}_{u, i}^2}.
    \end{aligned}
\end{equation*}
\end{lem}
\begin{proof}
Recalling that $\mathcal{P}$ is the ideal CVR estimation error in~\eqref{eq:ideal}, $q_{u,i}$ is the propensity score, $\hat{o}_{u,i}$ is the CTR estimate, $\varepsilon_{u,i}$ is the CVR estimation error in~\eqref{eq:cvrerrormeasure}. We have
\begin{equation*}
    \begin{aligned}
    \begin{aligned} \operatorname{Bias}\left[\mathcal{R}_\mathrm{DR}^\mathrm{err}\right] 
    &=\left|\mathbb{E}_O\left[\mathcal{R}_\mathrm{DR}^\mathrm{err}\right]-\mathcal{P}\right| \\ 
    &=\left| \frac{1}{|\mathcal{D}|} \sum_{(u, i) \in \mathcal{D}} \mathbb{E}_O\left[\hat{\varepsilon}_{u,i}+\frac{o_{u, i}\left(\varepsilon_{u,i}-\hat{\varepsilon}_{u,i}\right)}{\hat{o}_{u,i}}\right]-\varepsilon_{u,i}\right|\\ 
    &=\left|\frac{1}{|\mathcal{D}|} \sum_{(u, i) \in \mathcal{D}}\left[\hat{\varepsilon}_{u,i}+\frac{q_{u,i}\left(\varepsilon_{u,i}-\hat{\varepsilon}_{u,i}\right)}{\hat{o}_{u,i}}-\varepsilon_{u,i}\right]\right| \\ &=\frac{1}{|\mathcal{D}|}\left|\sum_{(u, i) \in D} \frac{q_{u,i}-\hat{o}_{u,i}}{\hat{o}_{u,i}}\left(\varepsilon_{u,i}-\hat{\varepsilon}_{u,i}\right)\right|, \end{aligned}
    \end{aligned}
\end{equation*}

\begin{equation*}
    \begin{aligned}
     \mathbb{V}_O\left[\mathcal{R}_\mathrm{DR}^\mathrm{err}\right] 
     &=\frac{1}{|\mathcal{D}|^2} \sum_{(u, i) \in \mathcal{D}} \mathbb{V}_O\left[\hat{\varepsilon}_{u,i}+\frac{o_{u, i}\left(\varepsilon_{u,i}-\hat{\varepsilon}_{u,i}\right)}{\hat{o}_{u,i}}\right] \\ &=\frac{1}{|\mathcal{D}|^2} \sum_{(u, i) \in \mathcal{D}} \mathbb{V}_O\left[o_{u, i}\right] \cdot\left(\frac{\varepsilon_{u,i}-\hat{\varepsilon}_{u,i}}{\hat{o}_{u,i}}\right)^2\\ &=\frac{1}{|\mathcal{D}|^2} \sum_{(u, i) \in \mathcal{D}} \frac{q_{u,i}\left(1-q_{u,i}\right)}{\hat{o}_{u,i}^2}\left(\hat{\varepsilon}_{u,i}-\varepsilon_{u,i}\right)^2. 
    \end{aligned}
\end{equation*}
\end{proof}

\begin{thm}[$\mathcal{R}_\mathrm{IPS}$ handles IEB]
Given accurate propensity score estimation, \ie $\hat{o}_{u,i}=q_{u,i}$, we have $\mathcal{R}_\mathrm{IPS} = \mathcal{P}$.
\end{thm}
\begin{proof}
Recalling that $o_{u,i}$ is the click indicator, $\hat{o}_{u,i}$ is the CTR estimate, $q_{u,i}$ is the actual propensity score. We have
\begin{subequations}\label{eq:ips_proof}
    \begin{align}
    \mathcal{R}_\mathrm{IPS} 
    &= \mathbb{E}_{(u,i)\in\mathcal{D}}\left[\frac{o_{u, i} \delta\left(r_{u, i}, \hat{r}_{u, i}\right)}{\hat{o}_{u, i}}\right] \notag\\
    &= \frac{|\mathcal{O}|}{|\mathcal{D}|}\mathbb{E}_{(u,i)\in\mathcal{O}}\left[\frac{ \delta\left(r_{u, i}, \hat{r}_{u, i}\right)}{\hat{o}_{u, i}}\right] \notag\\
    &= \frac{|\mathcal{O}|}{|\mathcal{D}|}\int\frac{ \delta\left(r_{u, i}, \hat{r}_{u, i}\right)}{\hat{o}_{u, i}} \mathbb{P}(u,i\mid O=1)\, d(u,i) \notag\\
    &= \frac{|\mathcal{O}|}{|\mathcal{D}|}\int\frac{ \delta\left(r_{u, i}, \hat{r}_{u,i}\right)}{\mathbb{P}(O=1\mid u,i)} \mathbb{P}(u,i\mid O=1)\, d(u,i) \label{eq:importance1}\\
    &= \mathbb{P}(O=1)\int\frac{ \delta\left(r_{u, i}, \hat{r}_{u,i}\right)}{\mathbb{P}(O=1\mid u,i)} \mathbb{P}(u,i\mid O=1)\, d(u,i) \\
    &= \int\delta\left(r_{u, i}, \hat{r}_{u, i}\right) \mathbb{P}(u,i)\, d(u,i) = \mathcal{P}\label{eq:importance2},
    \end{align}
\end{subequations}
where \eqref{eq:importance1} holds because we have $\hat{o}_{u,i}=q_{u,i}$; \eqref{eq:importance2} is derived by decomposing $\mathbb{P}(u,i\mid O=1)$ with the Bayesian formula.
\end{proof}

\begin{cor}[$\mathcal{R}_\mathrm{IPS}$ v.s. importance sampling]
$\mathcal{R}_\mathrm{IPS}$ is typically an importance sampling, which computes the ideal expectation over the exposure space $\mathcal{D}$ using samples from the click space $\mathcal{O}$.
\end{cor}
\begin{proof}
Using Bayes' theorem, \eqref{eq:importance1} could be reformulated as
\begin{equation}
\begin{aligned}
\int \delta\left(r_{u, i}, \hat{r}_{u,i}\right) \frac{\mathbb{P}(u,i)}{\mathbb{P}(u,i\mid O=1)} \mathbb{P}(u,i\mid O=1)\, d(u,i),
\end{aligned}
\end{equation}
which is a importance sampling to estimate the expectation of $\delta(r_{u,i}, \hat{r}_{u,i})$ in $\mathbb{P}(u,i)$, based on the samples from $\mathbb{P}(u,i\mid O=1)$. The importance weight is $\mathbb{P}(O=1)/\mathbb{P}(O=1\mid u,i)$.
\end{proof}

\begin{table*}
\centering
\caption{Performance on AUC, NDCG@K, and F1@K on the unbiased test set of Coat, Music and KuaiRec.}
\label{table:rating}
\renewcommand\arraystretch{1.3}
\begin{tabular}{l|ccc|ccc|ccc}
\hline
 & \multicolumn{3}{c|}{COAT} & \multicolumn{3}{c|}{MUSIC} & \multicolumn{3}{c}{KuaiRec} \\
Method & AUC & N@5 & F1@5 & AUC & N@5 & F1@5 & AUC & N@50 & F1@50 \\
\hline
Naive & 0.680$_{\pm 0.006}$ & 0.616$_{\pm 0.011}$ & 0.470$_{\pm 0.006}$ & 0.651$_{\pm 0.005}$ & 0.626$_{\pm 0.001}$ & 0.300$_{\pm 0.001}$ & 0.741$_{\pm 0.003}$ & 0.724$_{\pm 0.003}$ & 0.566$_{\pm 0.002}$ \\
ESMM & 0.686$_{\pm 0.004}$ & 0.638$_{\pm 0.005}$ & 0.485$_{\pm 0.008}$ & 0.601$_{\pm 0.002}$ & 0.665$_{\pm 0.001}$ & \textbf{0.328}$_{\pm 0.001}$ & 0.721$_{\pm 0.006}$ & 0.764$_{\pm 0.003}$ & 0.576$_{\pm 0.004}$ \\
MTL-EIB & 0.604$_{\pm 0.012}$ & 0.540$_{\pm 0.007}$ & 0.419$_{\pm 0.008}$ & 0.664$_{\pm 0.002}$ & 0.637$_{\pm 0.003}$ & 0.319$_{\pm 0.001}$ & 0.660$_{\pm 0.002}$ & 0.624$_{\pm 0.003}$ & 0.529$_{\pm 0.002}$\\
MTL-IMP & 0.713$_{\pm 0.003}$ & 0.613$_{\pm 0.011}$ & 0.462$_{\pm 0.004}$ & 0.627$_{\pm 0.003}$ & 0.661$_{\pm 0.003}$ & 0.328$_{\pm 0.002}$ & 0.735$_{\pm 0.006}$ & 0.719$_{\pm 0.007}$ & 0.573$_{\pm 0.006}$\\
MTL-IPS & 0.711$_{\pm 0.005}$ & 0.604$_{\pm 0.008}$ & 0.463$_{\pm 0.009}$ & 0.651$_{\pm 0.002}$ & \underline{0.667}$_{\pm 0.001}$ & 0.331$_{\pm 0.002}$ & 0.748$_{\pm 0.003}$ & 0.738$_{\pm 0.008}$ & 0.579$_{\pm 0.003}$ \\
MTL-DR & 0.719$_{\pm 0.006}$ & 0.634$_{\pm 0.009}$ & 0.480$_{\pm 0.007}$ & 0.686$_{\pm 0.001}$ & 0.660$_{\pm 0.003}$ & 0.323$_{\pm 0.002}$ & 0.752$_{\pm 0.001}$ & 0.767$_{\pm 0.012}$ & 0.581$_{\pm 0.003}$ \\\hline
AS-IPS & 0.712$_{\pm 0.008}$ & 0.627$_{\pm 0.010}$ & 0.470$_{\pm 0.007}$ & 0.661$_{\pm 0.003}$ & 0.641$_{\pm 0.004}$ & 0.322$_{\pm 0.003}$ & 0.746$_{\pm 0.009}$ & 0.733$_{\pm 0.004}$ & 0.585$_{\pm 0.006}$ \\
CVIB & 0.718$_{\pm 0.004}$ & 0.640$_{\pm 0.007}$ & 0.486$_{\pm 0.008}$ & 0.685$_{\pm 0.001}$ & 0.645$_{\pm 0.003}$ & 0.315$_{\pm 0.001}$ & 0.758$_{\pm 0.001}$ & 0.752$_{\pm 0.001}$ & 0.575$_{\pm 0.001}$ \\
DR-MSE & 0.715$_{\pm 0.001}$ & 0.630$_{\pm 0.009}$ & 0.480$_{\pm 0.006}$ & 0.685$_{\pm 0.001}$ & 0.648$_{\pm 0.004}$ & 0.316$_{\pm 0.002}$ & 0.779$_{\pm 0.003}$ & 0.773$_{\pm 0.004}$ & 0.589$_{\pm 0.001}$ \\
S-DR & 0.719$_{\pm 0.006}$ & 0.631$_{\pm 0.008}$ & 0.475$_{\pm 0.006}$ & \underline{0.687}$_{\pm 0.001}$ & 0.650$_{\pm 0.001}$ & 0.316$_{\pm 0.001}$ & 0.764$_{\pm 0.003}$ & \underline{0.791}$_{\pm 0.003}$ & 0.595$_{\pm 0.002}$ \\
\rowcolor{gray!10} ESCM${^2}$-IPS & \underline{0.721}$_{\pm 0.005}$ & \textbf{0.645}$_{\pm 0.009}$ & \textbf{0.490}$_{\pm 0.005}$ & 0.653$_{\pm 0.003}$ & 0.653$_{\pm 0.002}$ & 0.322$_{\pm 0.002}$ & \underline{0.779}$_{\pm 0.001}$ & 0.767$_{\pm 0.003}$ & \underline{0.592}$_{\pm 0.002}$ \\
\rowcolor{gray!10} ESCM${^2}$-DR & \textbf{0.730}$_{\pm 0.009}$ & \underline{0.642}$_{\pm 0.010}$ & \underline{0.489}$_{\pm 0.010}$ & \textbf{0.688}$_{\pm 0.002}$ & \textbf{0.669}$_{\pm 0.002}$ & \underline{0.326}$_{\pm 0.002}$ & \textbf{0.788}$_{\pm 0.001}$ & \textbf{0.796}$_{\pm 0.004}$ & \textbf{0.606}$_{\pm 0.002}$ \\
\hline
\end{tabular}
\begin{tablenotes}
\item[1] 1. Bold indicates the best performance. The underline marks the best performance across all baseline models except ESCM$^2$. 
\end{tablenotes}
\end{table*}

\setcounter{thm}{5}
\begin{thm}[$\mathcal{R}_\mathrm{IPS}$ handles PIP]
Suppose $\hat{r}^\mathrm{IPS}_{u,i}$ is the CVR estimate that optimizes $\mathcal{R}_\mathrm{IPS}$, $\mathbb{P}\left(r_{u, i}=1 \mid d o\left(o_{u, i}=1\right)\right)$ is the counterfactual conversion rate assuming the user clicked the item. For all samples in the exposure space, $\mathcal{R}_\mathrm{IPS}$ encourages:
\begin{equation*}
    \hat{r}_{u,i}^\mathrm{IPS}\rightarrow\mathbb{P}\left(r_{u, i}=1 \mid d o\left(o_{u, i}=1\right)\right).
\end{equation*}
\end{thm}
\begin{proof}

We start by dividing the samples in the exposure space into $K$ clusters $\{L_k\}_{k=1}^\mathrm{K}$.
Since the characteristics of samples in each cluster $L_k$ are similar, it is reasonable to assume that conditional exchangeability~\cite{peal_2009} holds in each cluster.
Formally, within the cluster $L_k$, the empirical distribution of clicks is independent of the distributions of counterfactual CVR estimation errors:
\begin{equation}
    \label{eq:exchange}
    \{\delta^{(1)}_k, \delta^{(0)}_k\}\upmodels O_k,
\end{equation}
where $\delta^{(1)}_k$ is the distribution of counterfactual CVR errors in $L_k$ assuming all users clicked the items, \ie $o_{u, i} = 1$ holds for all $(u,i)\in L_k$, $\delta^{(0)}_k$ is that given all users unclicked the items, \ie $o_{u, i} = 0$ holds for all $(u,i)\in L_k$. 

Then, based on the law of iterated expectations we have
\begin{equation}\label{eq:ips_reform}
    \mathcal{R}_{\mathrm{IPS}}
    =\mathbb{E}_{(u,i)\in\mathcal{D}}\left[\frac{o_{u, i}\varepsilon_{u,i}}{\hat{o}_{u, i}}\right]
    =\mathbb{E}_{k}\left\{\mathbb{E}_{(u,i)\in L_k}\left[\frac{o_{u,i}}{\hat{o}_{u,i}} \varepsilon_{u,i}\right]\right\},
\end{equation}
and within the cluster $L_k$ we have:
\begin{subequations}
\begin{align}
    \mathbb{E}_{(u,i)\in L_k}\left[\frac{o_{u,i}}{\hat{o}_{u,i}} \varepsilon_{u,i}\right]
    &=\mathbb{E}_{(u,i)\in L_k}\left[\frac{o_{u,i}}{\hat{o}_{u,i}} \varepsilon_{u,i}^{(1)}\right]\label{eq:exchange_ips_1}\\
    &=\mathbb{E}_{(u,i)\in L_k}\left[\frac{o_{u,i}}{\hat{o}_{u,i}}\right]\mathbb{E}_{(u,i)\in L_k}\left[\varepsilon_{u,i}^{(1)}\right]\label{eq:exchange_ips_2}\\
    &=\mathbb{E}_{(u,i)\in L_k}\left[\varepsilon_{u,i}^{(1)}\right]\triangleq \Delta_k,\label{eq:exchange_ips_3}
\end{align} 
\end{subequations}
where \eqref{eq:exchange_ips_1} and \eqref{eq:exchange_ips_2} holds as we have~\eqaref{eq:exchange}; $\Delta_k$ is the expectation of $\delta_k^{(1)}$. According to \eqaref{eq:exchange_ips_3} and \eqaref{eq:ips_reform} we have 
\begin{equation}
    \mathcal{R}_{\mathrm{IPS}}
    =\mathbb{E}_{k}\left[\Delta_k\right].
\end{equation}

Therefore, minimizing $\mathcal{R}_\mathrm{IPS}$ is equivalent to minimizing $\Delta_k$ in clusters $k=1,2,...,K$; and minimizing $\Delta_k$ encourages the convergence of the CVR estimate $\hat{r}_{u,i}^\mathrm{IPS}$ to the actual CVR assuming click takes place for all samples in the $k$-th cluster. The proof is completed.
\end{proof}

\begin{thm}[$\mathcal{R}_\mathrm{DR}$ handles IEB]
\label{thm:anti-ieb-dr}
Given either accurate propensity score estimation, \ie $\hat{o}_{u,i}=q_{u,i}$, or accurate error imputation, \ie $\hat{\varepsilon}_{u,i}=\varepsilon_{u,i}$, the equivalence $\mathcal{R}_\mathrm{DR} = \mathcal{P}$ holds.
\end{thm}
\begin{proof}
Recalling that $o_{u,i}$ is the click indicator, $\hat{o}_{u,i}$ is the CTR estimate, $q_{u,i}$ is the actual propensity score, $\hat{\varepsilon}_{u,i}$ is the imputed CVR error. 
First, given accurate imputation model, \ie $\hat{\varepsilon}_{u,i}=\varepsilon_{u,i}$ and $\hat{e}_{u,i}=0$, \eqref{eq:drreg} degrades to
\begin{equation*}
    \mathcal{R}_{\mathrm{DR}}^{\mathrm{err}}=\int\delta\left(r_{u, i}, \hat{r}_{u, i}\right) \mathbb{P}(u,i)\, d(u,i) = \mathcal{P}.
\end{equation*}

Second, given accurate propensity model, \ie $\hat{o}_{u,i}=q_{u,i}$, following \eqref{eq:ips_proof} we have
\begin{equation*}
\begin{aligned}
&\ \ \ \mathbb{E}_{(u, i) \in \mathcal{D}}\left[\frac{o_{u, i} \hat{e}_{u, i}\left(\phi_\mathrm{CVR},\phi_\mathrm{IMP}\right)}{\hat{o}_{u, i}}\right]\\
&=\mathbb{E}_{(u, i) \in \mathcal{D}}\left[\frac{o_{u, i} \hat{e}_{u, i}\left(\phi_\mathrm{CVR},\phi_\mathrm{IMP}\right)}{q_{u, i}}\right]\\
&=\int \hat{e}_{u,i} \mathbb{P}(u, i) d(u, i),
\end{aligned}
\end{equation*}
and \eqref{eq:drreg} can be simplified as
\begin{equation*}
\begin{aligned}
\mathcal{R}_{\mathrm{DR}}^{\mathrm{err}}&=\int \left(\hat{\varepsilon}_{u,i}+\hat{e}_{u,i}\right) \mathbb{P}(u, i) d(u, i)\\
\mathcal{R}_{\mathrm{DR}}^{\mathrm{err}}&=\int \left(\varepsilon_{u,i}\right) \mathbb{P}(u, i) d(u, i)=\mathcal{P}.
\end{aligned}
\end{equation*}
As such, either $\hat{o}_{u,i}=q_{u,i}$ or $\hat{\varepsilon}_{u,i}=\varepsilon_{u,i}$ makes $\mathcal{R}_{\mathrm{DR}}^{\mathrm{err}}=\mathcal{P}$, the proof is completed.
\end{proof}

\begin{thm}[$\mathcal{R}_\mathrm{DR}$ handles PIP]
\label{thm:anti-pip-dr}
Suppose $\hat{r}^\mathrm{DR}_{u,i}$ is the CVR estimate that optimizes $\mathcal{R}_\mathrm{DR}$, $\mathbb{P}\left(r_{u, i}=1 \mid d o\left(o_{u, i}=1\right)\right)$ is the counterfactual conversion rate assuming the user clicked the item. For all samples in the exposure space, $\mathcal{R}_\mathrm{DR}$ encourages:

\begin{equation*}
    \hat{r}_{u,i}^\mathrm{DR}\rightarrow\mathbb{P}\left(r_{u, i}=1 \mid d o\left(o_{u, i}=1\right)\right).
\end{equation*}
\end{thm}
\begin{proof}
Following the proof of \thmref{thm:anti-pip}, we start by dividing the samples in the exposure space into $K$ clusters $\{L_k\}_{k=1}^\mathrm{K}$, such that conditional ex-changeability holds in each $L_k$.

\begin{equation}
    \label{eq:exchange_dr}
    \{\hat{e}^{(0)}_k, \hat{e}^{(1)}_k\}\upmodels O_k,
\end{equation}
where $\hat{e}^{(1)}_k$ is the distribution of counterfactual imputation errors given $o_{u, i} = 1$ for all $(u,i)\in L_k$, $\hat{e}^{(0)}_k$ is that given $o_{u, i} = 0$ for all $(u,i)\in L_k$.

Then, based on the law of iterated expectations:
\begin{equation*}
\begin{aligned}
    \mathcal{R}_{\mathrm{DR}}^\mathrm{err}
    &=\mathbb{E}_{(u,i)\in\mathcal{D}}\left[\hat{\varepsilon}_{u,i}+\frac{o_{u, i}\hat{e}_{u,i}}{\hat{o}_{u, i}}\right]\\
    &=\mathbb{E}_{k}\left\{\mathbb{E}_{(u,i)\in L_k}\left[\hat{\varepsilon}_{u,i}+\frac{o_{u,i}}{\hat{o}_{u,i}} \hat{e}_{u,i}\right]\right\},
\end{aligned}
\end{equation*}
and within the group $L_k$, following \eqaref{eq:exchange_dr} we have:
\begin{equation*}
\begin{aligned}
    \mathbb{E}_{(u,i)\in L_k}\left[\frac{o_{u,i}}{\hat{o}_{u,i}} \hat{e}_{u,i}\right]
    &=\mathbb{E}_{(u,i)\in L_k}\left[\frac{o_{u,i}}{\hat{o}_{u,i}} \hat{e}_{u,i}^{(1)}\right]\\
    &=\mathbb{E}_{(u,i)\in L_k}\left[\frac{o_{u,i}}{\hat{o}_{u,i}}\right]\mathbb{E}_{(u,i)\in L_k}\left[\hat{e}_{u,i}^{(1)}\right]\\
    &=\mathbb{E}_{(u,i)\in L_k}\left[\hat{e}_{u,i}^{(1)}\right]\triangleq \Delta_k.
\end{aligned} 
\end{equation*}
Then, the DR regularizer can be reformulated as
\begin{equation*}
    \begin{aligned}
        \mathcal{R}_{\mathrm{DR}}^\mathrm{err}
    &=\mathbb{E}_{k}\left\{\mathbb{E}_{(u,i)\in L_k}\left[\hat{\varepsilon}_{u,i}+\frac{o_{u,i}}{\hat{o}_{u,i}} \hat{e}_{u,i}\right]\right\}\\
    &=\mathbb{E}_{k}\left\{\mathbb{E}_{(u,i)\in L_k}\left[\hat{\varepsilon}_{u,i}+\hat{e}_{u,i}^{(1)}\right]\right\}\\
    &=\mathbb{E}_{k}\left\{\mathbb{E}_{(u,i)\in L_k}\left[\Delta_k\right]\right\}.
    \end{aligned}
\end{equation*}

Therefore, minimizing $\mathcal{R}_\mathrm{DR}^\mathrm{err}$ is equivalent to minimizing $\Delta_k$ in clusters $k=1,...,K$. In particular, in the group $L_k$, $\Delta_k$ is the sum of imputed error and its counterfactual correction. As such, minimizing $\Delta_k$ encourages the CVR estimate $\hat{r}_{u,i}^\mathrm{DR}$ to converge to the actual CVR given click happens, for all samples in the $k$-th cluster.
\end{proof}

\subsection{Additional experimental results}

In this section, we explore the application of ESCM$^2$ to the task of rating prediction to demonstrate its versatility across different tasks. Rating prediction involves predicting whether the user assigns rating for items to infer their preference.
Likewise the CVR estimation task, the rating assignment is not random due to user preference: users are more likely to rate items that align with their interests and less likely to interact with items that do not. Thus, ratings are missing not at random, and this task is similarly challenged by sample selection bias, making it an ideal context for testing the efficacy of ESCM$^2$.

There are three datasets involved for this task, namely Coat, Music and KuaiRec, following~\cite{li2022stabilized}. Alongside the baseline methods established for CVR estimation, we also incorporate methods specifically designed for the rating prediction task, including AS-IPS~\cite{saito2020asymmetric}, CVIB~\cite{wang2020information}, DR-MSE~\cite{dai2022generalized} and S-DR~\cite{li2022stabilized}. The comparative results are present in Table~\ref{table:rating}, where ESCM$^2$ remains advantageous to outperform CVR baselines as well as exemplar baselines tailored for rating prediction.

\end{document}